\documentclass[journal]{IEEEtran}
\usepackage{times}
\usepackage{soul}
\usepackage{url}
\usepackage[hidelinks]{hyperref}
\usepackage[utf8]{inputenc}
\usepackage[small]{caption}
\usepackage{graphicx}
\usepackage{amsmath}
\usepackage{amsthm}
\usepackage{booktabs}
\usepackage[switch]{lineno}
\usepackage{amssymb,amsfonts}
\usepackage{enumitem}
\usepackage{multirow}
\usepackage{hyperref}
\hypersetup{
    colorlinks=true,
    linkcolor=blue,
    citecolor=green,
    urlcolor=blue
}
\usepackage{color, soul}
\newcommand{\justify}{\leftskip=0pt \rightskip=0pt plus 0cm}

\usepackage{subfigure}
\usepackage{subcaption}
\usepackage{algorithmic}
\usepackage[ruled,vlined,linesnumbered]{algorithm2e}
\usepackage[normalem]{ulem}
\useunder{\uline}{\ul}{}

\usepackage{tikz}
\usetikzlibrary{fadings}
\usetikzlibrary{mindmap,trees}
\usetikzlibrary{arrows,automata,shapes,positioning,shadows,trees}
\usetikzlibrary{shapes,snakes,shadows}
\usetikzlibrary{shapes.arrows}
\usetikzlibrary{calc,shapes, positioning}
\usetikzlibrary{decorations.text}
\usetikzlibrary{matrix,chains,positioning,decorations.pathreplacing,arrows}
\usetikzlibrary{bayesnet}
\usepackage[edges]{forest}
\usetikzlibrary{shadows.blur}
\usetikzlibrary{shapes.geometric}

\begin{document}

\title{Graph Condensation: A Survey}

\author{Xinyi~Gao, 
Junliang~Yu,
Tong~Chen, 
Guanhua~Ye,
Wentao~Zhang,
Hongzhi~Yin

\thanks{This work is supported by Australian Research Council under the streams of Future Fellowship (Grant No. FT210100624), Linkage Project (Grant No. LP230200892), Discovery Early Career Researcher Award (Grants No. DE230101033 and No. DE250100613), and Discovery Project (Grants No. DP240101108 and No. DP240101814).}
\thanks{Xinyi Gao, Junliang Yu, Tong Chen and Hongzhi Yin are affiliated with the School of Electrical Engineering and Computer Science, the University of Queensland, Brisbane, Australia.
Guanhua Ye is with the School of Computer Science, Beijing University of Posts and Telecommunications, Beijing, China. 
Wentao Zhang is with the Center for Machine Learning Research, Peking University, Beijing, China.}
\thanks{Corresponding author: Hongzhi Yin (e-mail: h.yin1@uq.edu.au).}
}

\markboth{Journal of \LaTeX\ Class Files,~Vol.~14, No.~8, August~2015}%
{Xinyi Gao \MakeLowercase{\textit{et al.}}: Bare Demo of IEEEtran.cls for IEEE Journals}

\IEEEtitleabstractindextext{
\begin{abstract} 
\justify{The rapid growth of graph data poses significant challenges in storage, transmission, and particularly the training of graph neural networks (GNNs). To address these challenges, graph condensation (GC) has emerged as an innovative solution. 
GC focuses on synthesizing a compact yet highly representative graph, enabling GNNs trained on it to achieve performance comparable to those trained on the original large graph.
The notable efficacy of GC and its broad prospects have garnered significant attention and spurred extensive research. 
This survey paper provides an up-to-date and systematic overview of GC, organizing existing research into five categories aligned with critical GC evaluation criteria: effectiveness, generalization, efficiency, fairness, and robustness. 
To facilitate an in-depth and comprehensive understanding of GC, this paper examines various methods under each category and thoroughly discusses two essential components within GC: optimization strategies and condensed graph generation. 
We also empirically compare and analyze representative GC methods with diverse optimization strategies based on the five proposed GC evaluation criteria.
Finally, we explore the applications of GC in various fields, outline the related open-source libraries, and highlight the present challenges and novel insights, with the aim of promoting advancements in future research. The related resources can be found at \href{https://github.com/XYGaoG/Graph-Condensation-Papers}{https://github.com/XYGaoG/Graph-Condensation-Papers}.
}
\end{abstract}

}

\maketitle
\IEEEdisplaynontitleabstractindextext
\IEEEpeerreviewmaketitle

\section{Introduction}
\IEEEPARstart{G}{raph} data is extensively utilized across a diverse range of domains, owing to its capability to represent complex structural relationships among various entities in the real world \cite{wu2022graph,gao2023accelerating,gao2023semantic}. Notable applications include, but are not limited to, social networks \cite{li2018influence,jung2012irie}, chemical molecular structures \cite{guo2022graph}, transportation systems \cite{rahmani2023graph}, and recommender systems \cite{zheng2023automl,yu2023self}.
However, the exponential growth of data volume in these applications makes the management and processing of large-scale graph data a complex and challenging task. A particularly demanding aspect is the computational requirements associated with training graph neural networks (GNNs) on large-scale graph datasets \cite{hamilton2017inductive}. These challenges are more pronounced in scenarios that require training multiple GNNs, such as hyper-parameter optimization \cite{ding_faster_2022}, continual learning \cite{yuan2023continual} and neural architecture search \cite{oloulade2021graph}, etc. This underscores the growing importance of developing efficient and effective methodologies for processing large-scale graph data. 

\begin{figure}[t]
\centering
\includegraphics[width=0.48\textwidth]{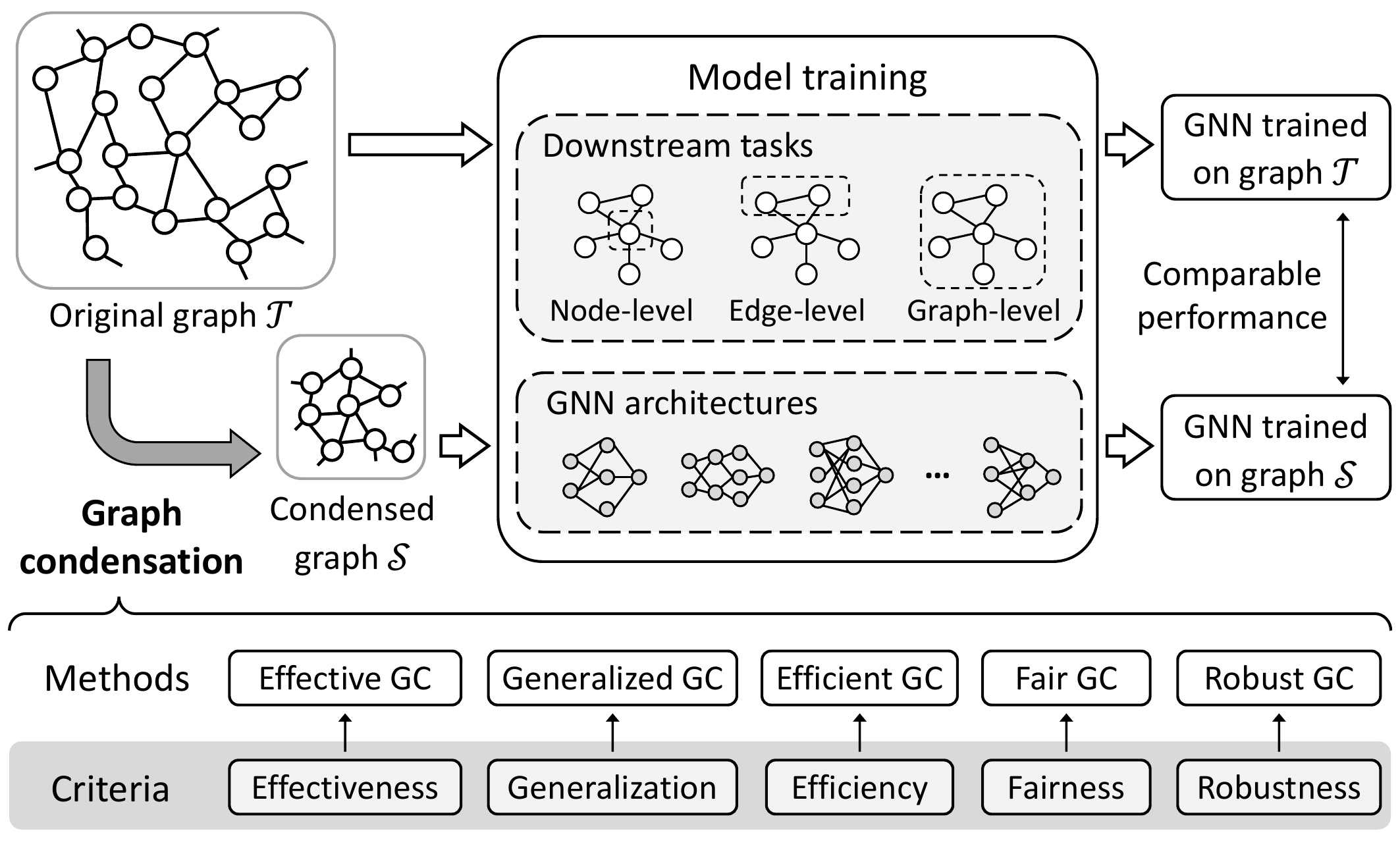}
\captionsetup{skip=10pt}
\caption{An overview for graph condensation. Graph condensation aims to generate small informative graphs such that the models trained on these graphs have similar downstream task performance to those trained on the original graphs. GC methods can be categorised into five classes aligned with critical GC evaluation criteria.} 
\label{fig:main}
\end{figure}

Early attempts to accelerate GNN training predominantly focused on model-centric methods, which aim to design advanced model structures to alleviate the heavy computational burden associated with the message-passing paradigm \cite{MPNN} in GNNs. 
Specifically, sampling-based methods \cite{hamilton2017inductive, DBLP:conf/iclr/ChenMX18, DBLP:conf/iclr/ZengZSKP20} attempt to minimize the number of nodes involved in message-passing by selectively sampling node neighbors. 
Linear aggregation methods \cite{wu2019simplifying,zhang2022nafs,zhang2022graph} propagate messages as a pre-processing step to reduce the online training computation.
Despite achieving more elaborate model architectures, these model-centric methods suffer from poor generalization and fail to ensure optimal performance across diverse graph datasets and downstream tasks \cite{ma2024acceleration}.
In light of these limitations, research attention has shifted towards data-centric methods \cite{yang2023data}, and graph reduction techniques have been proposed to accelerate model training by reducing graph size.
Appropriate simplification of graphs can not only expedite graph algorithms but also enhance the efficiency of storage, transformation, and retrieval processes for diverse graph analysis tasks.
Early explorations into graph reduction primarily focused on graph sparsification \cite{spielman2008graph} and coarsening \cite{loukas2018spectrally} techniques. These methods aim to reduce graph size by eliminating redundant edges and merging similar nodes. 
While effective in maintaining the essential characteristics of large graphs, these approaches have notable limitations.
They predominantly rely on heuristic methods, such as the largest principal eigenvalues \cite{loukas2018spectrally} or pairwise distances \cite{bravo2019unifying}, which still result in poor generalization across different downstream tasks \cite{hashemi2024comprehensive}.
Consequently, the emphasis of current research is on developing more generalized and adaptable graph reduction techniques, and the leading development is graph condensation (GC) (a.k.a. graph distillation).
As shown in Figure \ref{fig:main}, GC \cite{jin_graph_2022} aims to synthesize a small but highly informative graph dataset, encompassing both node feature and topology structure, to effectively represent the large original graph. 
Through task-driven optimization, models trained on these small condensed datasets can achieve performance comparable to those trained on the large original graphs. 
For instance, GCond \cite{jin_graph_2022} can condense the Flickr graph dataset to just 0.1\% of its original size, while some certain GNNs models trained on this small graph manage to retain 99.8\% of the original test accuracy.
Given the notable efficacy of GC, it has soon been applied to a variety of applications \cite{gao_graph_2023,liu_cat_2023,ding_faster_2022} and has sparked a proliferation of follow-up research \cite{zheng_structure_free_2023,xu_kernel_2023}, further broadening the reach and deepening the impact of GC.

Although significant strides have been made to enhance the effectiveness of GC methods, their evaluation extends beyond mere effectiveness due to the data-centric focus of these methods. This necessitates a broader exploration of additional criteria for assessing the performance of GC. 
One critical aspect is the condensed graph's ability to generalize across multiple GNN architectures and various downstream tasks \cite{jin_graph_2022}. A GC method with superior generalization is particularly beneficial in scenarios that demand the training of heterogeneous models or the handling of diverse tasks \cite{yang_does_2023}. 
In addition, it is essential for the condensed graph to maintain unbiased representations compared to the original graph \cite{liu2023fairgraph,dong2023reliant}, which is particularly crucial in fields sensitive to bias, such as human resources \cite{bourmpoulias2023entity}, criminal justice \cite{lettieri2023knowledge}, and financial services \cite{wang2021review}. 
Moreover, a robust GC procedure can enhance the resilience of condensed graphs to noisy graph structures or features \cite{zhu2021deep} in real-world systems.
Beyond these considerations, the scalability of GC also emerges as a key factor in practical applications \cite{gao2023accelerating}. An efficient GC method enables rapid adaption in life-long graph learning scenarios, effectively handling the continuous growth and dynamic changes of graph data. 
These criteria, effectiveness, generalization, efficiency, fairness, and robustness, play a pivotal role in GC evaluation and simultaneously serve as guiding principles for research motivations. Thus, we categorize existing GC methods based on these evaluation criteria, to provide a structured taxonomy for a comprehensive field survey. This taxonomy can help researchers comprehensively understand the diverse motivations driving GC methods, thereby facilitating a clearer path toward achieving their specific objectives.

While recent surveys on related topics including dataset distillation \cite{geng_survey_2023,yu_dataset_2023,sachdeva_data_2023,lei_comprehensive_2024}, data-centric graph learning \cite{yang2023data,zheng_towards_2023,shabani2024comprehensive}, as well as graph neural network acceleration \cite{zhang2023survey,ma2024acceleration} have touched upon GC, they summarize GC as a peripheral application or future direction, lacking systematic updates on the latest research and a holistic view. Moreover, concurrent GC surveys \cite{hashemi2024comprehensive,xu2024survey} and subsequent GC benchmarks \cite{liu2024gcondenser,gong2024gc,sun2024gc} focus on detailed optimization strategies, missing high-level perspectives and analysis of design objectives and rationals. This hinders researchers from understanding the novel GC for varying contexts. Recognizing the rapid evolution and growing significance of GC, there is an urgent need for a systematic survey that encapsulates the latest advancements and motivations behind methodological designs.

The contributions of this survey are summarized as follows:
\begin{itemize}
\item We are the first\footnote{The initial version was released on January 22, 2024, and is available at \href{https://arxiv.org/abs/2401.11720}{https://arxiv.org/abs/2401.11720}.} to survey the literature on GC and introduce a systematic taxonomy for GC, categorizing existing methods into five distinct groups based on key criteria: effectiveness, generalization, efficiency, fairness and robustness. 
This taxonomy structures existing knowledge and captures the underlying motivations driving GC research.

\item We provide a comprehensive and up-to-date review of the latest advancements in GC. Additionally, we delve into two fundamental aspects: optimization strategies and condensed graph generation, facilitating an in-depth comprehension of GC techniques. 

\item Through rigorous experiments, we systematically compare and analyze diverse optimization strategies in GC based on the five proposed GC evaluation criteria, offering comprehensive guidelines and insights for the design and development of advanced GC methods.

\item We explore practical applications and available resources in GC, while also shedding light on current challenges and future directions in the field. This discussion aims to guide researchers who are seeking to advance GC further.
\end{itemize}

The structure of this survey is outlined as follows: Section 2 delves into the definition, criteria, and taxonomy of GC. 
Following this, Section 3 discusses the optimization strategies employed in GC methods. 
Section 4 provides a thorough investigation of each category of GC, as aligned with our taxonomy. 
In Section 5, we extensively discuss the methods involved in condensed graph generation, and Section 6 empirically compares the representative GC methods across diverse evaluation criteria.
Section 7 is dedicated to summarizing the various applications and resources of GC. 
Finally, we present the challenges and future directions in this field in Section 8.
\begin{figure}[t]
\centering
\includegraphics[width=0.46\textwidth]{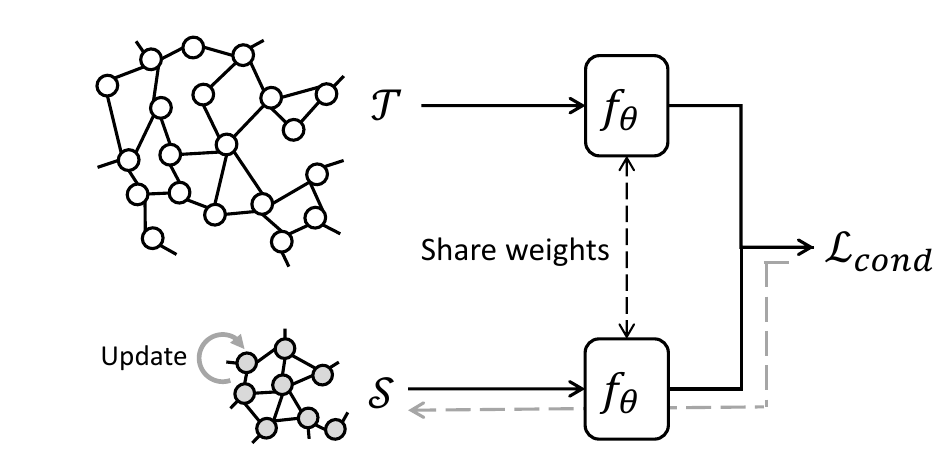}
\captionsetup{skip=10pt}
\caption{Graph condensation procedure. The original graph and condensed graph are encoded by the relay model $f_{\theta}$ and the condensed graph is optimized according to $\mathcal{L}_{cond}$.} 
\label{fig:GC}
\end{figure}

\tikzstyle{leaf}=[draw=black,
    rounded corners,minimum height=1em,
    fill=black!15,text opacity=1, align=center,
    fill opacity=.5,  text=black,align=left,font=\footnotesize,
    inner xsep=3pt,
    inner ysep=1pt,
    ]
\tikzstyle{middle}=[draw=black,
    rounded corners,minimum height=1em,
    fill=white!40,text opacity=1, align=center,
    fill opacity=.5,  text=black,align=center,font=\footnotesize,
    inner xsep=3pt,
    inner ysep=1pt,
    ]
    
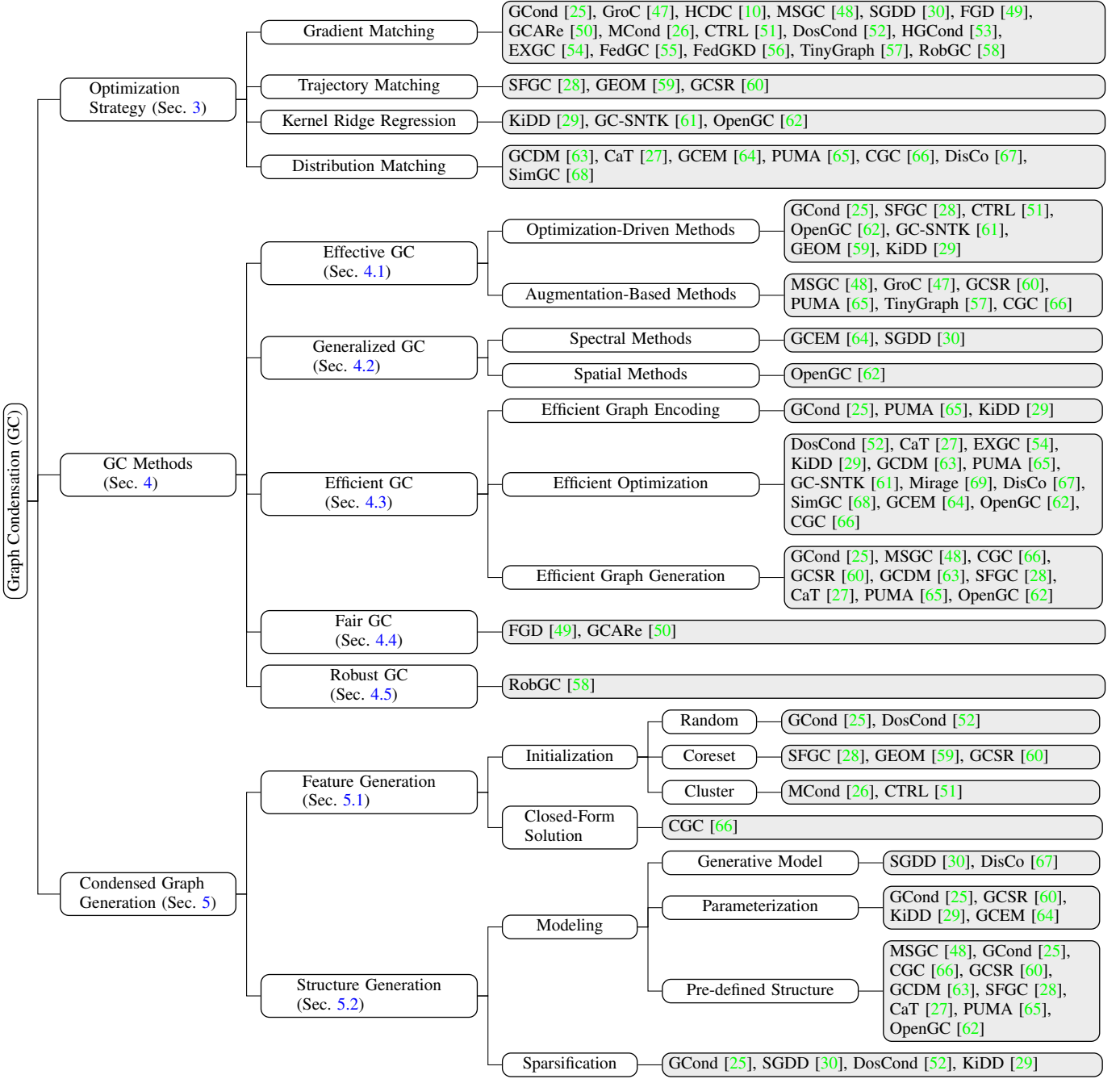
\begin{figure*}[ht]
\centering
\begin{forest}
  for tree={
    forked edges,
    grow=east,
    reversed=true,
    anchor=base west,
    parent anchor=east,
    child anchor=west,
    base=middle,
    font=\footnotesize,
    rectangle,
    line width=0.5pt,
    draw=black,
    rounded corners,align=left,
    minimum width=2em,
    s sep=5pt,
    inner xsep=3pt,
    inner ysep=1pt,
  },
  where level=1{text width=4.5em}{},
  where level=2{text width=6em,font=\footnotesize}{},
  where level=3{font=\footnotesize}{},
  where level=4{font=\footnotesize}{},
  where level=5{font=\footnotesize}{},
[{Graph Condensation (GC)}, middle, rotate=90, anchor=north, edge=black
    [Optimization \\ Strategy (Sec. \ref{secopt}), middle, edge=black, text width=7em
        [Gradient Matching, middle, text width=9em, edge=black
            [GCond \cite{jin_graph_2022}{,} GroC \cite{li_attend_2023}{,} HCDC \cite{ding_faster_2022}{,} MSGC \cite{gao_multiple_2023}{,} SGDD \cite{yang_does_2023}{,} FGD \cite{feng_fair_2023}{,} \\GCARe \cite{mao_gcare_2023}{,} MCond \cite{gao_graph_2023}{,} CTRL \cite{CTRL}{,} DosCond \cite{jin_condensing_2022}{,} HGCond \cite{gao2024heterogeneous}{,} \\EXGC \cite{fang2024exgc}{,} FedGC \cite{yan2024federated}{,} FedGKD \cite{pan_fedgkd_2023}{,} TinyGraph \cite{liu2024tinygraph}{,} RobGC \cite{gao2024robgc}, leaf, text width=28em, edge=black]
        ]
        [Trajectory Matching, middle, text width=9em, edge=black
            [SFGC \cite{zheng_structure_free_2023}{,} GEOM \cite{zhang2024navigating}{,} GCSR \cite{liu2024graph}, leaf, text width=28em, edge=black]
        ]
        [Kernel Ridge Regression, middle, text width=9em, edge=black
            [KiDD \cite{xu_kernel_2023}{,} GC-SNTK \cite{wang_fast_2023}{,} OpenGC \cite{gao2024graph}, leaf, text width=28em, edge=black]
        ]
        [Distribution Matching, middle, text width=9em, edge=black
            [GCDM \cite{liu_graph_2022}{,} CaT \cite{liu_cat_2023}{,} GDEM \cite{liu_graph_2023}{,} PUMA \cite{liu_puma_2023}{,} CGC \cite{gao2024rethinking}{,}  DisCo \cite{xiao2024disentangled}{,} \\SimGC \cite{xiao2024simple}, leaf, text width=28em, edge=black]
        ]
    ]
    [GC Methods \\ (Sec. \ref{secmet}), middle, edge=black, text width=7em
        [Effective GC \\(Sec. \ref{secmeteffe}), middle, edge=black,text width=9em
            [Optimization-Driven Methods, middle, text width=12em, edge=black
                [GCond \cite{jin_graph_2022}{,} SFGC \cite{zheng_structure_free_2023}{,} CTRL \cite{CTRL}{,} \\OpenGC \cite{gao2024graph}{,} GC-SNTK \cite{wang_fast_2023}{,} \\GEOM \cite{zhang2024navigating}{,} KiDD \cite{xu_kernel_2023}, leaf, text width=14em, edge=black]
            ]
            [Augmentation-Based Methods, middle, text width=12em, edge=black
                [MSGC \cite{gao_multiple_2023}{,} GroC \cite{li_attend_2023}{,} GCSR \cite{liu2024graph}{,} \\PUMA \cite{liu_puma_2023}{,} TinyGraph \cite{liu2024tinygraph}{,} CGC \cite{gao2024rethinking}, leaf, text width=14em, edge=black]
            ]
        ]
        [Generalized GC \\(Sec. \ref{secmetgene}), middle, edge=black, text width=9em
            [Spectral Methods, middle, text width=12em, edge=black
                [GDEM \cite{liu_graph_2023}{,} SGDD \cite{yang_does_2023}, leaf, text width=14em, edge=black]
            ]
            [Spatial Methods, middle, text width=12em, edge=black
                [OpenGC \cite{gao2024graph}, leaf, text width=14em, edge=black]
            ]
        ]
        [Efficient GC \\(Sec. \ref{secmeteffi}), middle, edge=black, text width=9em
            [Efficient Graph Encoding, middle, text width=12em, edge=black
                [GCond \cite{jin_graph_2022}{,} PUMA \cite{liu_puma_2023}{,} KiDD \cite{xu_kernel_2023}, leaf, text width=14em, edge=black]
            ]
            [Efficient Optimization, middle, text width=12em, edge=black
                [DosCond \cite{jin_condensing_2022}{,} CaT \cite{liu_cat_2023}{,} EXGC \cite{fang2024exgc}{,} \\KiDD \cite{xu_kernel_2023}{,}  GCDM \cite{liu_graph_2022}{,} PUMA \cite{liu_puma_2023}{,} \\GC-SNTK \cite{wang_fast_2023}{,} Mirage \cite{gupta_mirage_2023}{,} DisCo \cite{xiao2024disentangled}{,} \\SimGC \cite{xiao2024simple}{,} GDEM \cite{liu_graph_2023}{,} OpenGC \cite{gao2024graph}{,} \\CGC \cite{gao2024rethinking}, leaf, text width= 14em, edge=black]
            ]
            [Efficient Graph Generation, middle, text width=12em, edge=black
                [GCond \cite{jin_graph_2022}{,} MSGC \cite{gao_multiple_2023}{,} CGC \cite{gao2024rethinking}{,} \\GCSR \cite{liu2024graph}{,} GCDM \cite{liu_graph_2022}{,} SFGC \cite{zheng_structure_free_2023}{,} \\CaT \cite{liu_cat_2023}{,} PUMA \cite{liu_puma_2023}{,} OpenGC \cite{gao2024graph},leaf, text width= 14em, edge=black]
            ]
        ]
        [Fair GC \\(Sec. \ref{secmetfair}), middle, edge=black, text width=9em
            [FGD \cite{feng_fair_2023}{,} GCARe \cite{mao_gcare_2023}, leaf, text width=28em, edge=black]
        ]
        [Robust GC \\(Sec. \ref{secmetrobu}), middle, edge=black, text width=9em
            [RobGC \cite{gao2024robgc}, leaf, text width=28em, edge=black]
        ]
    ]
    [{Condensed Graph \\ Generation (Sec. \ref{secgene})}, middle, edge=black, text width=7em
        [{Feature Generation \\ (Sec. \ref{secgeneatt})}, middle, edge=black, text width=9em
            [Initialization, middle, edge=black, text width=6em
                [Random, middle, text width=4em, edge=black
                    [GCond \cite{jin_graph_2022}{,} DosCond \cite{jin_condensing_2022}, leaf, text width=14em, edge=black]
                ]
                [Coreset, middle, text width=4em, edge=black
                    [SFGC \cite{zheng_structure_free_2023}{,} GEOM \cite{zhang2024navigating}{,} GCSR \cite{liu2024graph},leaf, text width= 14em, edge=black]
                ]
                [Cluster, middle, text width=4em, edge=black
                    [MCond \cite{gao_graph_2023}{,} CTRL \cite{CTRL},leaf, text width= 14em, edge=black]
                ]
            ]
            [Closed-Form \\Solution, middle, text width=6em, edge=black
                [CGC \cite{gao2024rethinking}, leaf, text width=20em, edge=black]
            ]
        ]
        [{Structure Generation \\ (Sec. \ref{secgenestru})}, middle, edge=black, text width=9em
            [Modeling, middle, text width=6em, edge=black
                [Generative Model, middle, text width=9em, edge=black
                    [SGDD \cite{yang_does_2023}{,} DisCo \cite{xiao2024disentangled}, leaf, text width=9em, edge=black]
                ]
                [Parameterization, middle, text width=9em, edge=black
                    [GCond \cite{jin_graph_2022}{,} GCSR \cite{liu2024graph}{,} \\KiDD \cite{xu_kernel_2023}{,} GDEM \cite{liu_graph_2023},leaf, text width= 9em, edge=black]
                ]
                [Pre-defined Structure, middle, text width=9em, edge=black
                    [MSGC \cite{gao_multiple_2023}{,} GCond \cite{jin_graph_2022}{,} \\CGC \cite{gao2024rethinking}{,} GCSR \cite{liu2024graph}{,} \\GCDM \cite{liu_graph_2022}{,} SFGC \cite{zheng_structure_free_2023}{,} \\CaT \cite{liu_cat_2023}{,} PUMA \cite{liu_puma_2023}{,} \\OpenGC \cite{gao2024graph},leaf, text width= 9em, edge=black]
                ] 
            ]
            [Sparsification, middle, text width=6em, edge=black
                [GCond \cite{jin_graph_2022}{,} SGDD \cite{yang_does_2023}{,} DosCond \cite{jin_condensing_2022}{,} KiDD \cite{xu_kernel_2023}, leaf, text width= 20em, edge=black]
            ] 
        ]
    ]
] 
\end{forest}
\caption{The taxonomy of graph condensation.}
\label{fig:tax}
\end{figure*}

\section{Definition and Taxonomy}
\label{sec2}
\subsection{Preliminaries}

\noindent\textbf{Graph dataset.} A large-scale graph dataset can be represented as $\mathcal{T} =\left(\mathcal{V}, \mathcal{E}\right)$, which contains $|\mathcal{V}| = N$ nodes and $|\mathcal{E}|=M$ edges \footnote{We unify definitions for node-level and graph-level graph datasets.}.
Its $d$-dimensional node feature matrix, adjacency matrix, and task specific label are denoted as ${\bf X}\in{\mathbb{R}^{N\times d}}$,  ${\bf A}\in \mathbb{R}^{N\times N}$ and $\mathbf{Y}$, respectively. The entry in the adjacency matrix ${\bf A}_{i,j}>0$ denotes an observed edge from node $i$ to $j$, and ${\bf A}_{i,j}=0$ otherwise.

\noindent\textbf{Graph neural network.} GNNs leverage the graph topological information ${\bf A}$ and node features ${\bf X}$ to learn node representation via message-passing \cite{MPNN}, formulating GNN layers by aggregation and transformation functions. The $l^{th}$ layer of GNNs is formulated as:
\begin{equation}
\begin{aligned}
    & \mathbf{m}_v^{(l)} \leftarrow \operatorname{aggregate}  
    \left( \mathbf{h}_v^{(l-1)},\left\{\mathbf{h}_u^{(l-1)} \mid u \in \mathcal{N}(v)\right\}\right), \\
    & \mathbf{h}_v^{(l)} \leftarrow \operatorname{transform} 
    \left( \mathbf{m}_v^{(l)},\mathbf{h}_v^{(l-1)} \right),
\end{aligned}
\end{equation}
where $\mathbf{h}_v^{(l)}$ represents the representation of node $v$ in the $l^{th}$ layer, while $\mathbf{m}_v^{(l)}$ denotes the message for node $v$. 
The function $\operatorname{aggregate}(\cdot, \cdot)$ aggregates the neighboring nodes of node $v$ (i.e., $\mathcal{N}(v)$) to compute the message $\mathbf{m}_v^{(l)}$, which is subsequently updated by the transformation function $\operatorname{transform}(\cdot, \cdot)$.
Node representations are initialized as node features, and the final representations after $L$ GNN layers are utilized for various downstream tasks.

\noindent\textbf{Graph condensation.} Graph condensation aims to find a small condensed graph dataset $\mathcal{S} = \left(\mathcal{V}', \mathcal{E}'\right)$ with $|\mathcal{V}'| = N'$ and $N' \ll N$, enabling more efficient GNN training while maintaining performance comparable to GNNs trained on the original graph $\mathcal{T}$.
The adjacency matrix, node feature matrix, and task-specific label of $\mathcal{S}$ are denoted as ${\mathbf{A}'}$, $\mathbf{X}'$ and $\mathbf{Y}'$, respectively. 
The compression rate (a.k.a. condensation ratio) $r$ is defined as $\frac{N'}{N}$. 
To facilitate the connection between original graph $\mathcal{T}$ and condensed graph $\mathcal{S}$, a relay graph model $f_{\theta}\left(\cdot\right)$, parameterized by ${\theta}$, is employed in the optimization process for encoding both graphs as shown in Figure \ref{fig:GC}. Then, the graph condensation is formulated as an optimization problem:
\begin{equation}
\label{definition}
\mathcal{S} = \arg\min_{\mathcal{S}}\mathcal{L}_{cond}\left(f_{\theta}\left({\mathcal{T}}\right), f_{\theta}\left({\mathcal{S}}\right) \right),
\end{equation}
where $\mathcal{L}_{cond}$ is the optimization objective for graph condensation, which varies in format and will be elaborated upon in Section \ref{secopt}.

\subsection{Criteria for Graph Condensation}
The primary metric for assessing GC methods involves evaluating the accuracy of GNN models trained on condensed graphs as compared to those trained on the original graphs, providing a direct measure of information preservation. To achieve a comprehensive evaluation and gain a holistic understanding of method performance, it is essential to incorporate additional criteria.
Specifically, we summarize the key criteria of GC as follows:
\begin{itemize}
\item \textbf{Effectiveness} assesses the accuracy of a specified GNN architecture trained on the condensed graph under various compression rates \cite{jin_graph_2022}. It measures the extent to which the condensed graph preserves the essential characteristics of the original graph.
\item \textbf{Generalization} measures the overall accuracy of different GNN architectures trained on the condensed graph across various downstream tasks \cite{yang_does_2023}. It reflects the condensed graph's adaptability and utility in diverse contexts.
\item \textbf{Efficiency} assesses the time required to condense a graph \cite{jin_condensing_2022}, focusing on the speed of condensed graph generation and the method's feasibility for time-sensitive scenarios.
\item \textbf{Fairness} evaluates the disparity in model performance between GNNs trained on condensed graphs and those trained on original graphs across different demographic groups\cite{feng_fair_2023}, aiming to prevent bias amplification in the condensed graph and ensure equitable model performance.
\item \textbf{Robustness} measures the quality of condensed graphs generated from the noisy original graph. 
It assesses the ability of GC methods to identify and preserve the vital characteristics of the original graph despite the presence of noise in structures or features.

\end{itemize}
Given these criteria, an ideal GC method should generate a high-fidelity condensed graph dataset at low cost, enabling models with different GNN architectures to achieve results comparable and unbiased to those obtained on the original graph on a wide range of downstream tasks.

\begin{figure}[t]
    \centering
    \subfigure[Split by GC criteria.]
    {
    \begin{minipage}[b]{.47\linewidth}
        \centering
        \includegraphics[width=\linewidth]{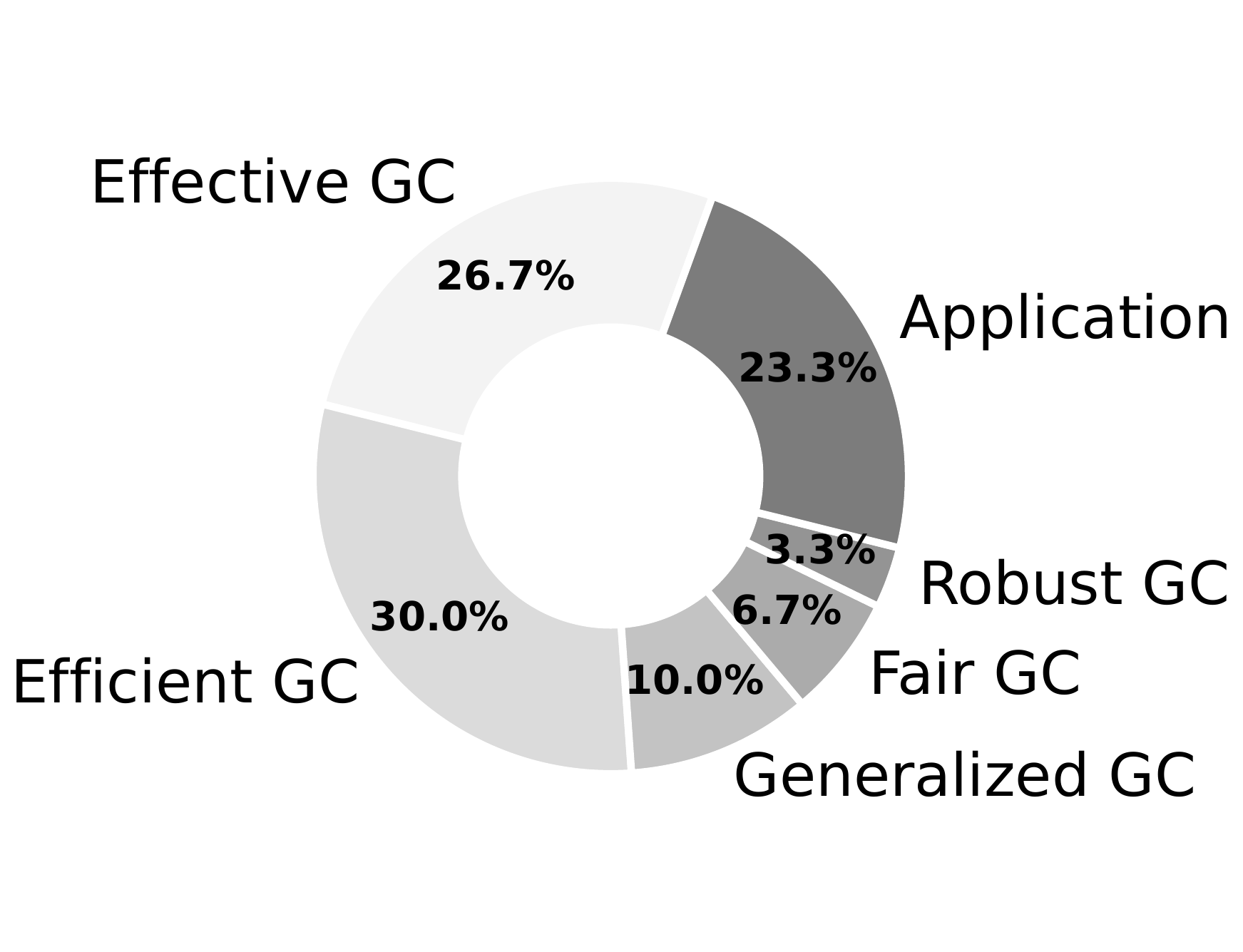}
        \label{pie1}
    \end{minipage}
    }
    \subfigure[Split by optimization objectives.]
    {
    \begin{minipage}[b]{.45\linewidth}
        \centering
        \includegraphics[width=\linewidth]{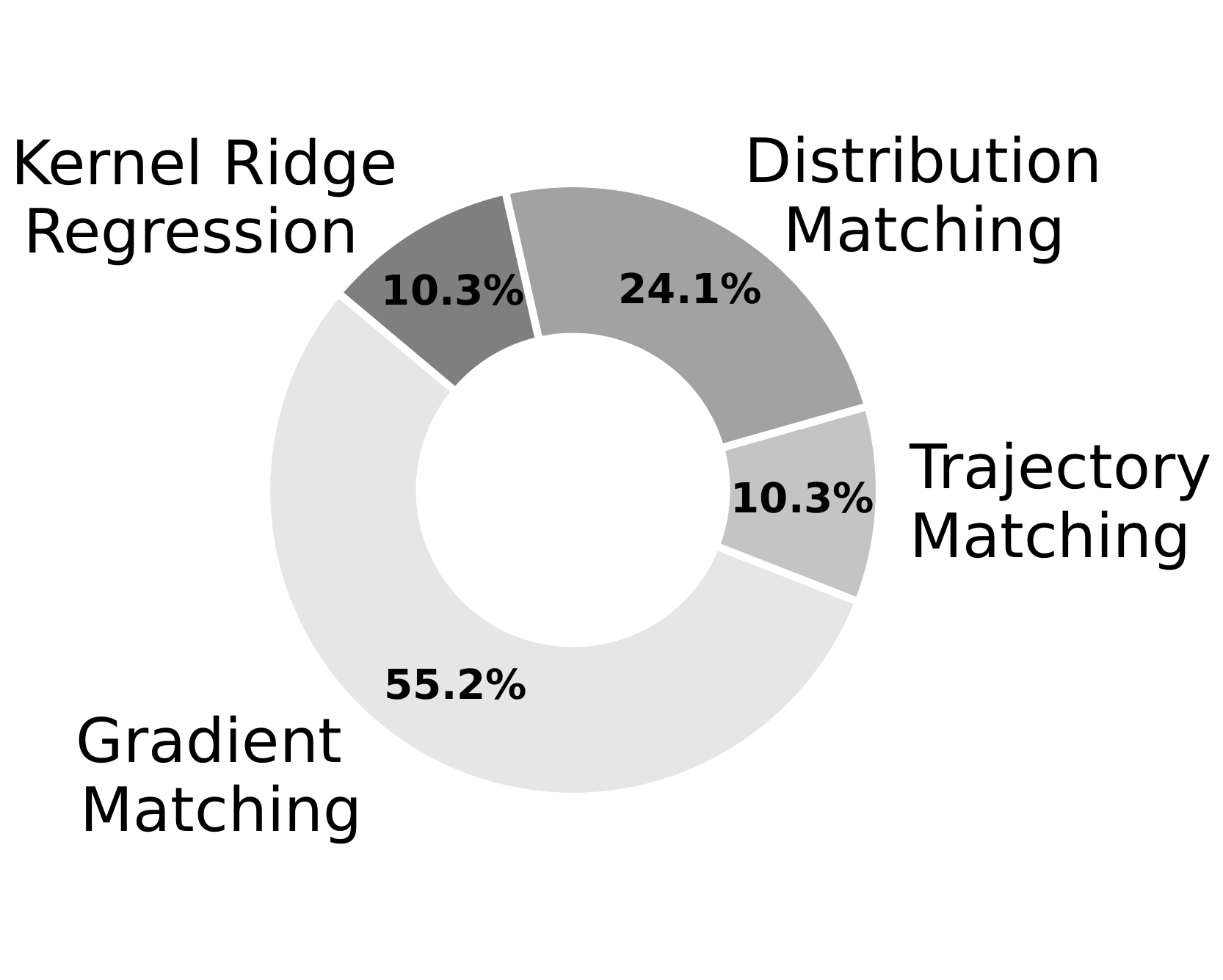}
        \label{pie2}
    \end{minipage} 
    }
    
    \caption{The research focuses of graph condensation literature.}
    \label{figpie}
\end{figure}

\subsection{Taxonomy}
The above criteria not only serve as indicators of utility for GC but also act as the objectives that steer the extensive research efforts in this field. To streamline the understanding of GC, we class GC methods into five categories that align with the above criteria: effective GC, generalized GC, efficient GC, fair GC, and robust GC, as shown in Figure \ref{fig:tax}. The paper distribution for each category is show in Figure \ref{figpie} (a).

\subsubsection{\textbf{Effective Graph Condensation}}
\begin{figure}[ht]
\centering
\includegraphics[width=0.4\textwidth]{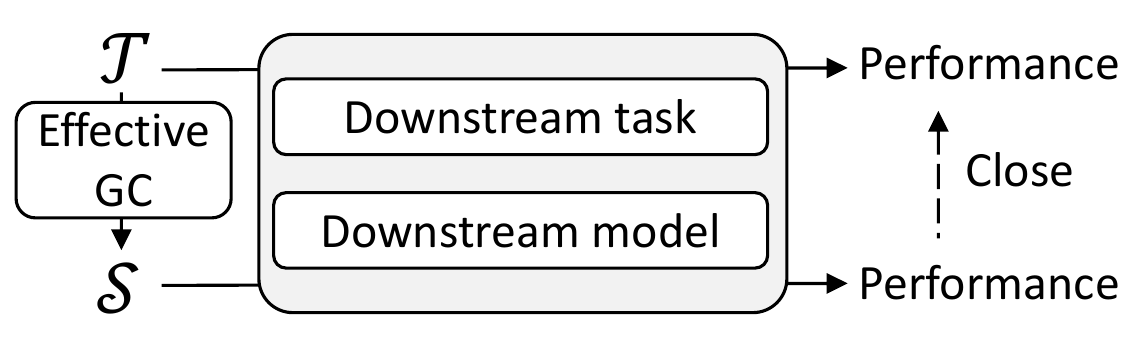}
\captionsetup{skip=15pt}
\caption{Effective graph condensation (GC). The condensed graph enables GNNs to be trained with performance comparable to GNNs trained on the original graph.} 
\label{fig:GC1}
\end{figure}

The primary objective of GC focuses on optimizing the task-specific quality of the condensed graph. As illustrated in Figure \ref{fig:GC1}, the condensed graph produced by effective GC methods enables GNNs to achieve performance that closely approximates those trained on the original graph, especially in targeted downstream tasks.
To systematically study these methods, we further divide this category into two distinct classes: optimization-driven methods and augmentation-based methods, facilitating an in-depth understanding of techniques in effective GC.

\vspace{8pt}

\subsubsection{\textbf{Generalized Graph Condensation}}

\begin{figure}[ht]
\centering
\includegraphics[width=0.4\textwidth]{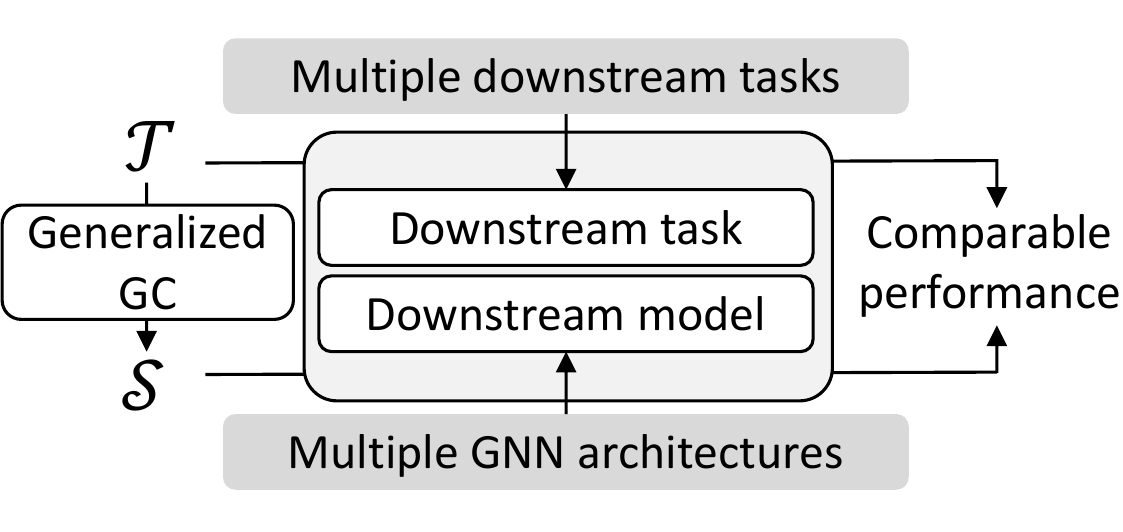}
\captionsetup{skip=15pt}
\caption{Generalized graph condensation (GC). The condensed graph facilitates the training of various GNN architectures and supports various downstream tasks.} 
\label{fig:GC2}
\end{figure}

The objective of GC extends beyond merely training specific models for designated tasks. It also involves enabling the application of various models across a spectrum of tasks, as depicted in Figure \ref{fig:GC2}. A condensed graph with good generalization capabilities avoids the repeated generation of task-specific or model-specific condensed graphs, thereby significantly enhancing its utility in various applications, such as open-world graph learning \cite{wu2020openwgl}, neural architecture search \cite{oloulade2021graph} and hyper-parameter optimization \cite{bergstra2011algorithms}.
The primary challenge of generalization is to capture the key information in both the structural and feature representations during condensation.
Consequently, we study existing generalized GC from two primary aspects: the spectral and spatial techniques for better information preservation during condensation.
 
\vspace{8pt}

\subsubsection{\textbf{Efficient Graph Condensation}}
\begin{figure}[ht]
\centering
\includegraphics[width=0.5\textwidth]{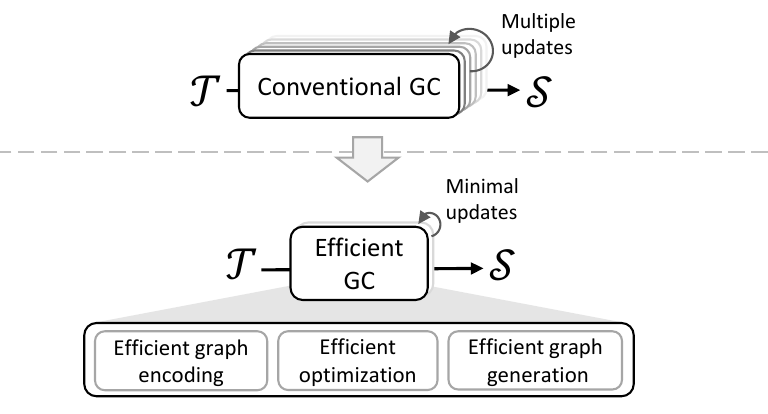}
\captionsetup{skip=15pt}
\caption{Efficient graph condensation (GC) contributes to accelerating the condensation procedure.} 
\label{fig:GC4}
\end{figure}

An efficient condensing procedure is essential for rapid adaptation in life-long graph learning scenarios. However, GC often suffers from sophisticated optimization processes and slow convergence, resulting in a time-consuming procedure. This particularly deteriorates in life-long learning contexts, where regular updates to the condensed graph dataset are a common requirement, as depicted in Figure \ref{fig:GC4}. To address these challenges, efficient GC methods have been developed to accelerate the condensation process across all components within GC. To thoroughly explore the mechanisms underlying efficient GC, we break down the conventional GC procedure into three stages: graph encoding, optimization, and graph generation. These components align with the input, processing, and output stages of GC, respectively. By enhancing the efficiency of each stage, efficient GC methods significantly reduce the time required for the entire GC process.

\vspace{8pt}

\subsubsection{\textbf{Fair Graph Condensation}}
\begin{figure}[ht]
\centering
\includegraphics[width=0.4\textwidth]{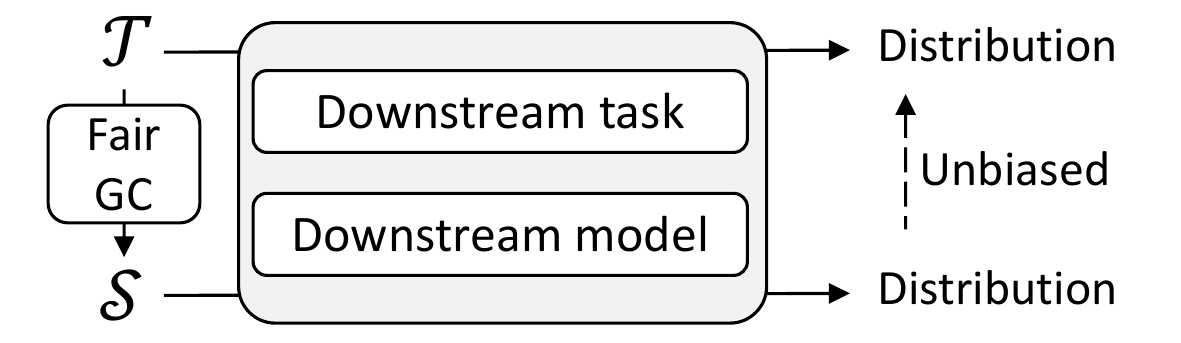}
\captionsetup{skip=15pt}
\caption{Fair graph condensation (GC) ensures that GNNs trained on condensed graphs produce results unbiased relative to those from GNNs trained on the original graph.} 
\label{fig:GC3}
\end{figure}

Balancing information during the condensation process represents a pivotal challenge in the domain of GC. The intensive compression of GC often leads to an amplification of data biases within the condensed graphs. Consequently, GNNs trained on such condensed graphs exhibit more pronounced fairness issues compared to those trained on original graphs.
In response to this, fair GC methods constrain the condensation procedure to mitigate bias in the condensed graph as shown in Figure \ref{fig:GC3}.
The approaches to improving fairness in GC encompass two key aspects: the first integrates the fairness-enhancing regularization into the GC objective function, effectively guiding the optimization process toward fairer outcomes. The second aspect is the development of advanced relay model, promoting unbiased representations in the condensed graphs.

\vspace{8pt}

\subsubsection{\textbf{Robust Graph Condensation}}
\begin{figure}[ht]
\centering
\includegraphics[width=0.4\textwidth]{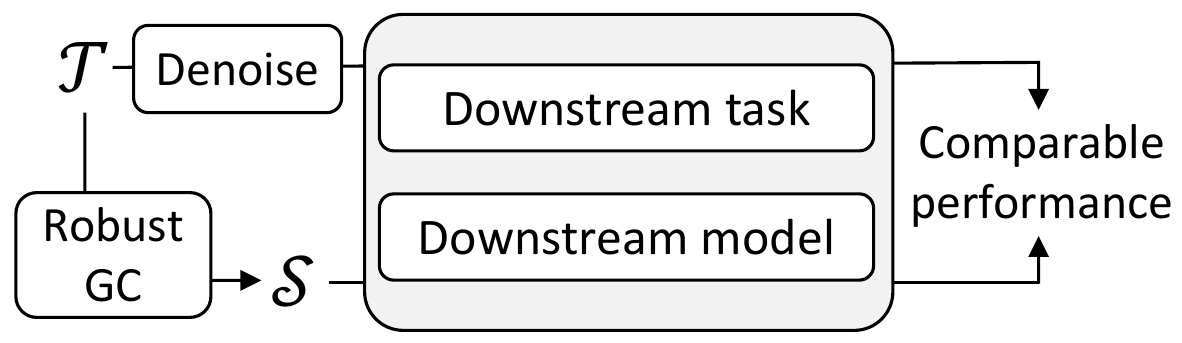}
\captionsetup{skip=15pt}
\caption{Robust graph condensation (GC) ensures that GNNs trained on condensed graphs produce comparable results relative to those from GNNs trained on the clean original graph.} 
\label{fig:GC5}
\end{figure}

A robust condensation procedure is significant for unskewed condensed graph generation and deployment in noisy real-world scenarios. 
Conventional GC indiscriminately emulates the original graph distributions by condensed graph, disregarding the intrinsic quality of the graph.
As a result, the condensed graph inherits the noisy distribution from the original graph, which is carried over to GNNs and compromises the model prediction accuracy.
To address this issue, Robust GC aims to filter out the noise within the original graph during the condensation procedure and extracts only the core, causal information to the condensed graph for effective GNN training, as depicted in Figure \ref{fig:GC5}. 

\vspace{8pt}

\section{Optimization Strategies}
\label{secopt}
As we delve deeper into the intricacies of GC, it is crucial to examine the optimization strategies that form a core module within the GC framework. We hereby introduce the various optimization strategies employed in GC, offering a high-level perspective on the condensation procedure.
Our discussion commences with the foundational principle of optimization objective (Eq. (\ref{eq_pmloss})), then moves onto detailed optimization strategies (Eq. (\ref{eq_gmloss1})-(\ref{eq_dmloss})).

The foundational principle of the optimization strategies posits that the performance of a model trained on the condensed data should align with that of a model trained on the original data. 
To this end, the losses for data $\mathcal{T}$ and $\mathcal{S}$ in relation to the parameter ${\theta}$ of the relay model are initially defined as:
\begin{equation}
\label{loss}
\begin{split}
&\mathcal{L}^{\mathcal{T}}\left(\theta\right) = \ell \left(f_{\theta}\left({\mathcal{T}}\right), \mathbf{Y}\right),\\
&\mathcal{L}^{\mathcal{S}}\left(\theta\right) = \ell \left(f_{\theta}\left({\mathcal{S}}\right), \mathbf{Y}'\right),  
\end{split}
\end{equation}
where $\ell$ is the task-specific objective such as cross-entropy. Then the objective of GC can be formulated as the following bi-level problem:
\begin{equation}
\label{eq_pmloss}
\min_{\mathcal{S}} \mathcal{L}^{\mathcal{T}} \left(\theta^{\mathcal{S}}\right) 
\:\: \text{s.t.}  \:\: \theta^{\mathcal{S}}   = \arg\min_{\mathcal{\theta}} \mathcal{L}^{\mathcal{S}} \left(\theta\right).
\end{equation}

Addressing the objective outlined in Eq. (\ref{eq_pmloss}) necessitates the incorporation of the task-specific objective $\ell$ and entails resolving a nested loop optimization, which includes unrolling the entire training trajectory of the relay model optimization (i.e., inner loop) and then updates the condensed graph in the outer loop.
A general workflow for this bi-level optimization can be formulated as Algorithm \ref{algorithm} and this process can be prohibitively expensive \cite{zhao2020dataset}. 
To alleviate the complexity of the optimization process, various approximation methods have been proposed as shown in Figure \ref{figpie} (b), including gradient matching, trajectory matching, kernel ridge regression, and distribution matching.

\begin{algorithm}[t]
\textbf{Input:} Original graph $\mathcal{T}$, relay model $f_{\theta}$, iterations \{$T_1$, $T_2$\}.\\
\textbf{Output:} Condensed graph $\mathcal{S}$.\\
Initialise $\mathcal{S}$ and $f_{\theta}$.\\
\For({\qquad\qquad\qquad\qquad\hspace{0pt} {$\rhd$ Outer loop}}){$t_1$= 1 to $T_1$}
{
Update $\mathcal{S}$ according to $\mathcal{L}^{\mathcal{T}}$.\\
\For({\qquad\qquad\qquad\hspace{4pt} {$\rhd$ Inner loop}}){$t_2$= 1 to $T_2$}
{
Optimize $f_{\theta}$ according to $\mathcal{L}^{\mathcal{S}}$.\\
}
}
\textbf{Return:} Condensed graph $\mathcal{S}$.\\
\caption{The bi-level optimization workflow.}
\label{algorithm}
\end{algorithm}

\begin{table*}[h]
\centering
\caption{The comparison of the time complexity for diverse optimization objectives. The process in the condensing procedure varies for different GC objectives. Gradient matching includes the gradient calculation. Kernel ridge regression includes the computation of regression solution. Distribution matching includes class representation calculation.}
{\begin{tabular}{l|l|l|l|l|l|l}
\toprule
 \multirow{2}{*}{\textbf{Strategy}}      & \multirow{2}{*}{\textbf{Pre-processing}} & \multicolumn{5}{c}{\textbf{Condensing procedure}}   \\ \cline{3-7} 
       &       & \textbf{Forward} & \textbf{Process}      & \textbf{Loss} & \textbf{Update} $\mathcal{S}$& \textbf{Update} $f_{\theta}$\\ \midrule
Gradient matching  & $\mathcal{O} \left(LMd \right)$ & $\mathcal{O} \left( \left(N+N' \right)dh \right)$ & $\mathcal{O} \left(dh \right)$           & $\mathcal{O} \left(dh \right)$  & $\mathcal{O} \left(N'd \right)$           & $\mathcal{O} \left(qdh \right)$      \\
Trajectory matching   & $\mathcal{O} \left(Z(LMd+Ndh) \right)$& $\mathcal{O} \left(qN'dh \right)$    & N/A& $\mathcal{O} \left(dh \right) $ & $\mathcal{O} \left(N'd \right)$           & $\mathcal{O} \left(qdh \right)$\\
Kernel ridge regression & $\mathcal{O} \left(LMd\right)$          & $\mathcal{O} \left( \left(N+N' \right)dh \right)$ & $\mathcal{O} \left(N'^3+NCh \right)$ & $\mathcal{O} \left(Ch \right)$  & $\mathcal{O} \left(N'd \right)$  & N/A \\ 
Distribution matching   & $\mathcal{O} \left(LMd \right)$ & $\mathcal{O} \left( \left(N+N' \right)dh \right)$ & $\mathcal{O} \left( \left(N+N' \right)h \right)$      & $\mathcal{O} \left(Ch \right) $ & $\mathcal{O} \left(N'd \right)$           & N/A\\
\bottomrule
\end{tabular}}
\label{tab_com}
\end{table*}

\noindent\textbf{Gradient matching.} 
Gradient matching is initially proposed by Zhao et al. \cite{zhao2020dataset} and has become the dominant optimization strategy in GC.
Specifically, it formulates the condensation objective in Eq. (\ref{eq_pmloss}) as matching the optimized parameters of the models trained on two datasets:
\begin{equation}
\label{eq_gmloss1}
\min_{\mathcal{S}} \mathbb{E}_{\theta_0 \sim {\Theta}} [\mathcal{D}\left({\theta}^{\mathcal{T}}, {\theta}^{\mathcal{S}}\right)],
\end{equation}
where $\theta_0$ is the initialization of ${\theta}^{\mathcal{T}}$ and ${\theta}^{\mathcal{S}}$. ${\Theta}$ is a specific distribution for relay model initialization. The expectation on $\theta_0$ aims to improve the robustness of ${\mathcal{S}}$ to different parameter initialization \cite{lei_comprehensive_2024}. $\mathcal{D}\left(\cdot,\cdot\right)$ is the distance measurement. 
The bi-level objective in Eq. (\ref{eq_pmloss}) is approximated by matching the model gradients at each training step $t$. In this way, the training trajectory on condensed data can mimic that on the original data, i.e., the models trained on these two datasets converge to similar solutions. The optimization objective $\mathcal{L}_{cond}$ in Eq. (\ref{definition}) is defined as:
\begin{equation}
\label{eq_gmloss2}
\begin{split}
& \mathcal{L}_{cond} =  \mathbb{E}_{\theta_0 \sim {\Theta}} \left [\sum_{t=1}^{T} \mathcal{D}\left(\nabla_{\theta}\mathcal{L}^{\mathcal{T}}\left({\theta_t}\right), \nabla_{\theta}\mathcal{L}^{\mathcal{S}}\left({\theta_t}\right)\right)\right] \,\, \\
& \text{s.t.} \,\, {\theta}_{t+1} = \operatorname{opt}\left(\mathcal{L}^\mathcal{S}\left({\theta_t}\right)\right),
\end{split}
\end{equation}
where $\operatorname{opt}\left(\cdot\right)$ is the model parameter optimizer and the parameter of relay model is updated only on $\mathcal{S}$.

\noindent\textbf{Trajectory matching.}
Gradient matching primarily aligns with single-step gradients, yet it may accumulate errors when the relay model is iteratively updated using condensed data across multiple steps. To mitigate this problem, Cazenavette et al. \cite{cazenavette2022dataset} propose a multi-step matching approach to approximate Eq. (\ref{eq_gmloss1}) known as trajectory matching.

Trajectory matching trains two separate relay models on condensed data and original data, respectively. These two relay models are trained for distinct steps starting from the same initialization $\theta'_t$, which is sampled from the original data training checkpoints ${\Theta'}$. 
Then, trajectory matching aims to minimize the discrepancy between the final points of these two training trajectories: 
\begin{equation}
\label{eq_tmloss}
\begin{split}
& \mathcal{L}_{cond} =  \mathbb{E}_{\theta'_t \sim {\Theta'}} \left[ \mathcal{D}\left({\theta}^{\mathcal{T}}_{t+k}, {\theta}^{\mathcal{S}}_{t+l}\right)\right] \\
& \text{s.t.} \,\, {\theta}^{\mathcal{T}}_{t+1} = \operatorname{opt}\left(\mathcal{L}^\mathcal{T}\left({\theta^{\mathcal{T}}_t}\right) \right), \\
& \,\,\,\,\,\,\,\,\, {\theta}^{\mathcal{S}}_{t+1} = \operatorname{opt}\left(\mathcal{L}^\mathcal{S}\left({\theta^{\mathcal{S}}_t}\right)\right),
\end{split}
\end{equation}
where $k$ and $l$ are hyper-parameters controlling the update steps. ${\theta}^{\mathcal{S}}_{t}$ and ${\theta}^{\mathcal{T}}_{t}$ denote the parameters of relay models trained on $\mathcal{T}$ and $\mathcal{S}$ at step $t$ respectively. Although trajectory matching introduces a tri-level optimization process, the training phase on the original data can be pre-processed. Consequently, the online condensation procedure remains a bi-level optimization process.

\noindent\textbf{Kernel ridge regression.}
Gradient matching and trajectory matching approximately optimize the objective Eq. (\ref{eq_gmloss1}) by matching one-step gradient or multi-step trajectory. However, the complex bi-level optimization and inexact solution inevitably lead to slow convergence and performance drop. 
In light of these challenges, Nguyen et al. \cite{nguyen2020dataset} transform the original task into a regression problem by substituting the neural network with kernel ridge regression (KRR). This approach enables the direct resolution of the inner loop optimization in Eq. (\ref{eq_pmloss}) using a closed-form solution. Consequently, the complex nested optimization is simplified into a first-order process by considering the condensed data as the support set and the original data as the target set.
The optimization problem described in Eq. (\ref{eq_pmloss}) is reformulated into a regression task: 
\begin{equation}
\label{eq_krrloss}
   \mathcal{L}_{cond} =  \frac{1}{2} \lVert {\bf Y} - K_{{\mathcal{T}}{\mathcal{S}}}\left(K_{{\mathcal{S}}{\mathcal{S}}} + \lambda I\right)^{-1} {\bf Y}' \rVert^2,
\end{equation}
where $K_{{\mathcal{T}}{\mathcal{S}}}$ is the kernel matrix of ${\mathcal{T}}$ and ${\mathcal{S}}$, and $K_{{\mathcal{S}}{\mathcal{S}}}$ is the kernel matrix of ${\mathcal{S}}$. It is important to note that matrix inversion is solely dependent on the number of condensed samples, making it computationally efficient.

\noindent\textbf{Distribution matching.}
All aforementioned optimization strategies stem from the objective of aligning model performance as outlined in Eq. (\ref{eq_pmloss}), inevitably introducing the relay model training procedure within the optimization process.
To avoid this issue, Zhao et al. \cite{zhao2023dataset} propose an alternative optimization principle known as distribution matching, which focuses on matching the feature distributions of condensed and original data.
The objective of distribution matching is to generate condensed data with a feature distribution closely approximating that of the original data. Accordingly, the optimization objective of distribution matching is formulated as:
\begin{equation}
\label{eq_dmloss}
\mathcal{L}_{cond} =  \mathbb{E}_{\theta_0 \sim {\Theta}} \left[\mathcal{D}\left(f_{\theta}\left({\mathcal{T}}\right), f_{\theta}\left({\mathcal{S}}\right)\right)\right].
\end{equation}
To address the discrepancy in sample quantities between the original and condensed dataset, the distance measurement $\mathcal{D}\left(\cdot,\cdot \right)$ is specifically designed for class-wise comparison, necessitating the use of class labels in distribution matching.

\begin{table*}[t]
\centering
\caption{Summary of existing graph condensation research, organized chronologically by publication date. 
In optimization strategy: ``GM'' represents gradient matching, ``DM'' refers to distribution matching, ``TM'' is short for trajectory matching, 
``KRR'' is kernel ridge regression, and ``CTP'' denotes computation tree preservation.
In relay model: ``GNN'' is graph neural network, ``GNTK'' refers to graph neural tangent kernel and ``SD'' is short for spectral decomposition.
In downstream task: ``NC'' represents node classification, ``GC'' refers to graph classification, ``AD'' is anomaly detection and ``LP'' is short for link prediction. The ``Enhanced Module'' identifies the specific target for improvements within the GC framework.
}
\label{tab:sum}
\renewcommand{\arraystretch}{1.5}
\resizebox{\textwidth}{!}
{
\begin{tabular}{l|l|c|llc|ll|c|l}
\toprule
\multirow{2}{*}{\textbf{Method}} & \multirow{2}{*}{\textbf{Category}} & \textbf{Input} & \multicolumn{3}{c|}{\textbf{Optimization Strategy}} & \multicolumn{2}{c|}{\textbf{Output}}    & \multirow{2}{*}{\textbf{\begin{tabular}[c]{@{}c@{}}Downstream\\ Task\end{tabular}}} & \multirow{2}{*}{\textbf{\begin{tabular}[c]{@{}c@{}}Enhanced\\ Module\end{tabular}}} \\ \cline{3-8}
                                 &                                      & Preproc.       & Objective         & Model         & Bi-level        & Structure construction & Sparsification &                                                                                     &                                                                                     \\

\midrule
GCond \cite{jin_graph_2022} &Effectiveness  &  & GM & GNN & \checkmark & Generative model & Threshold  & NC & N/A \\ \hline
DosCond \cite{jin_condensing_2022} & Efficiency &  & GM & GNN &  & Parameterization & Gumbel softmax  & GC & Optimization \\ \hline
GCDM \cite{liu_graph_2022} & Efficiency &  & DM & GNN & \checkmark & Generative model & Threshold  & NC & Objective \\ \hline
HCDC \cite{ding_faster_2022} & Application  &  & GM & GNN & \checkmark & Generative model & Threshold  & NC & Condensed graph \\ \hline
MSGC \cite{gao_multiple_2023} & Effectiveness & \checkmark & GM & GNN & \checkmark & Pre-defined structure &N/A  & NC & Condensed graph \\ \hline 
SGDD \cite{yang_does_2023} & Generalization & \checkmark & GM & GNN & \checkmark & Generative model &Regularization  & NC, AD, LP & Objective \\ \hline
SFGC \cite{zheng_structure_free_2023} & Effectiveness & \checkmark & TM & GNN &  & Pre-defined structure &N/A  & NC & Objective \\ \hline
FGD \cite{feng_fair_2023} & Fairness &  & GM & GNN & \checkmark & Generative model &Threshold  & NC & Objective \\ \hline
GCARe \cite{mao_gcare_2023} & Fairness &  & GM & GNN & \checkmark & Generative model &Threshold  & NC & Relay model \\ \hline
MCond \cite{gao_graph_2023} & Application &  & GM & GNN & \checkmark & Generative model &Threshold  & NC & Objective \\ \hline
KiDD \cite{xu_kernel_2023} & Efficiency & \checkmark & KRR & GNTK &  & Parameterization &Gumbel softmax  & GC & Relay model \\ \hline
CaT \cite{liu_cat_2023} & Application &  & DM & GNN &  & Pre-defined structure &N/A  & NC & Optimization \\ \hline
CTRL \cite{CTRL} & Effectiveness  &  & GM & GNN & \checkmark & Generative model &Threshold  & NC, GC & Objective \\ \hline
FedGKD \cite{pan_fedgkd_2023} & Application &  & GM & GNN &  & Generative model &Gumbel softmax  & NC & Condensed graph \\ \hline
GDEM \cite{liu_graph_2023}& Generalization  & \checkmark & DM & SD &  & Parameterization &N/A  & NC &  Relay model\\ \hline
GC-SNTK \cite{wang_fast_2023} & Efficiency     &     & KRR & GNTK &     & Pre-defined structure & N/A       & NC     & Objective       \\ \hline
Mirage \cite{gupta_mirage_2023} & Efficiency & \checkmark & CTP & Free  & & Parameterization &N/A  & GC & Optimization \\ \hline
GroC \cite{li_attend_2023} & Effectiveness  &  & GM & GNN & \checkmark & Generative model &Threshold  & NC, GC & Objective \\ \hline
PUMA \cite{liu_puma_2023} & Application & \checkmark & DM & GNN &  & Pre-defined structure &N/A  & NC & Objective \\ \hline 
DisCo \cite{xiao2024disentangled} & Efficiency & \checkmark & DM & MLP &  & Generative model &N/A  & NC & Objective \\ \hline
GEOM \cite{zhang2024navigating}   & Effectiveness  & \checkmark & TM  & GNN  &   & Generative model      & Threshold & NC     & Objective       \\ \hline
HGCond \cite{gao2024heterogeneous} & Application    &     & GM  & GNN  & \checkmark & Generative model      & N/A       & NC     & Condensed graph \\ \hline
EXGC \cite{fang2024exgc}   & Efficiency     &     & GM  & GNN  & \checkmark & Generative model      & Threshold & NC, GC & Optimization    \\ \hline
OpenGC \cite{gao2024graph}  & Generalization & \checkmark & KRR & GNN  &     & Pre-defined structure & N/A       & NC     & Objective       \\ \hline
GCSR \cite{liu2024graph}   & Effectiveness  &     & TM  & GNN  & \checkmark & Parameterization      & N/A       & NC     & Objective       \\ \hline
SimGC \cite{xiao2024simple}  & Efficiency     & \checkmark & DM  & GNN  &     & Generative model      & Threshold & NC     & Optimization    \\ \hline
FedGC \cite{yan2024federated}  & Application    &     & GM  & GNN  & \checkmark & Generative model      & Threshold & NC     & Objective       \\ \hline
CGC \cite{gao2024rethinking}    & Efficiency     & \checkmark & DM  & GNN  &     & Parameterization      & Threshold & NC     & Optimization    \\ \hline
RobGC \cite{gao2024robgc}  & Robustness     & \checkmark & GM  & GNN  & \checkmark & Generative model      & Threshold & NC     & Condensed graph \\ \hline
TinyGraph \cite{liu2024tinygraph}  & Effectiveness     & & GM  & GNN  & \checkmark & Generative model      & Threshold & NC     & Condensed graph \\
\bottomrule
\end{tabular}}
\end{table*}

\noindent\textbf{Discussion.}
The gradient matching approach focuses on aligning short-range model parameters, while trajectory matching enhances data quality by matching long-range model trajectories. 
Both methods employ the bi-level optimization framework, requiring multiple forward and backward computations and thus incurring high costs and time consumption.
In contrast, KRR and distribution matching adopt the single-level optimization framework, leading to more efficient optimization processes. 
Although distribution matching uses the feature space as a matching proxy and bypasses relay model optimization, this strategy inherently confines it to class-wise comparisons. Consequently, this design restricts its adaptability to diverse task-specific objectives and its applicability to data lacking class labels.

For an intuitive comparison, we present the detailed time complexity of various optimization strategies in Table \ref{tab_com}. 
Time complexities for both the pre-processing and condensing phases are assessed separately and the condensing procedure is further divided into forward propagation, process execution, loss calculation, condensed graph updating, and relay model updating. 
Following GCond \cite{jin_graph_2022}, the relay model $f_{\theta}$ for all strategies is a $L$ layer SGC \cite{wu2019simplifying} incorporating a linear model with the hidden dimension denoted by $h$.
For the original graph, $N$, $M$, $d$, and $C$ are the number of nodes, edges, feature dimensions, and classes, respectively. 
The number of nodes in the condensed graph is represented by $N'$.
For a fair comparison, the pre-defined adjacency matrix \cite{zheng_structure_free_2023} is utilized across all methods. 
The pre-processing stage for gradient matching \cite{jin_graph_2022}, kernel ridge regression \cite{gao2024graph} and distribution matching \cite{sun2024gc} incorporates non-parametric graph convolution. 
Trajectory matching \cite{zheng_structure_free_2023} entails the pre-training of teacher models and the model quantity is denoted by $Z$.
The process execution stage involves different operations specific to each strategy.
Specifically, gradient matching entails calculating the gradient w.r.t the relay model parameters twice.
Kernel ridge regression calculates the closed-form solution, and distribution matching involves computing the class representation. 
As for the update of the relay model, gradient matching and trajectory matching necessitate updating the relay model on the condensed graph $q$ times to generate model parameters and trajectories in the inner loop, respectively.
Conversely, kernel ridge regression and distribution matching simplify the process by eliminating bi-level optimization, thereby obviating the need to update the relay model.
\section{Graph Condensation Methods}
\label{secmet}
In this section, we dive into the specifics of GC and conduct a systematic analysis of five distinct GC methodologies: effective GC, generalized GC, efficient GC, fair GC, and robust GC, exploring the techniques employed to achieve their respective objectives.
In addition to our proposed taxonomy, we also provide a detailed summary of GC methods, highlighting their distinct characteristics, in Table \ref{tab:sum}. This table offers a comprehensive comparison across several dimensions, including category, optimization strategy, condensed graph generation method, and downstream tasks.

\subsection{Effective Graph Condensation}
\label{secmeteffe}

As mentioned in Section \ref{sec2}, we further divide effective GC methods into two categories according to the enhanced components within the GC framework, including optimization-driven methods and augmentation-based methods.

\subsubsection{\textbf{Optimization-Driven Methods}}
Designing more advanced optimization strategies is often considered the critical driving force to achieve effective GC, thus steering many studies in this field. The first GC method, \textbf{GCond} \cite{jin_graph_2022}, employs gradient matching as the optimization strategy. However, the limitations associated with single-step gradient alignment constrain the overall performance of GC. Therefore, follow-up studies are proposed to explore and enhance the optimization process.

\noindent\textbf{Gradient matching.}
\textbf{CTRL} \cite{CTRL} argues that the distance measurement in gradient matching is coarse, which relies on cosine similarity and solely focuses on the directional aspect of gradients.
This fails to adequately capture the vector characteristics of gradients, thereby introducing biases in the gradient matching process. Therefore, CTRL proposes to refine the matching objective by adding the magnitudes distance. Furthermore, CTRL empirical discovers that gradient magnitude matching not only explicitly accounts for the vector properties of gradients, but also implicitly benefits the alignment of frequency distributions between condensed and original graphs. This provides another compelling rationale for the performance improvement.

\noindent\textbf{Trajectory matching.}
Despite the improvements from fine-grained gradient matching, CTRL still focuses on short-range matching.
This leads to the short-sight issue and fails to capture the holistic GNN learning behaviors, potentially compromising the quality of condensed graphs. 
To address this limitation, \textbf{SFGC} \cite{zheng_structure_free_2023} introduces trajectory matching in GC and proposes to align the long-term GNN learning behaviors between the original graph and condensed graph. 
As a result, SFGC enables a more comprehensive knowledge transfer to the condensed graph.
Furthermore, \textbf{GEOM} \cite{zhang2024navigating} identifies the restricted supervision signals in trajectory matching and argues that difficult nodes are key to the performance gap of GNNs trained on the condensed graph. To address this, GEOM evaluates difficult nodes through the principle of homophily \cite{wei2023clnode} and utilizes curriculum learning to adjust the size of the matching window for trajectories. This approach allows GEOM to systematically structure the condensation procedure to progress from simpler to more challenging trajectories, thereby incorporating more diverse supervision signals from the original graph.

\noindent\textbf{Kernel ridge regression.}
Beyond aligning gradients or trajectories during model training, kernel ridge regression is employed for precise solutions in bi-level optimization objectives.
For instance, \textbf{GC-SNTK} \cite{wang_fast_2023} discards the GNN and employs the kernel ridge regression, incorporating graph neural tangent kernels \cite{du2019graph} as the relay model.
This modification not only facilitates the first-order optimization of the condensed graph but also eliminates the requirement for multiple initializations of relay GNN, resulting in a more stable condensation procedure.
\textbf{KiDD} \cite{xu_kernel_2023} employs kernel ridge regression for graph classification tasks and further improves the method by removing the non-linear activation in graph neural tangent kernels. 
This approach simplifies the matrix multiplications in the iterative updates of condensed graphs and leads to substantial improvements in model accuracy compared to the gradient matching method.

\subsubsection{\textbf{Augmentation-Based Methods}}
Optimization-driven methods enhance the effectiveness of GC by emphasizing the optimization procedure. In contrast, other studies improve the performance by augmenting different modules in the GC framework.

\noindent\textbf{Structure augmentation.}
\textbf{MSGC} \cite{gao_multiple_2023} and \textbf{GCSR} \cite{liu2024graph} improve the GC from graph structure construction. 
Unlike many GC approaches which synthesize a single dense adjacency matrix, MSGC instead pre-defines multiple small-scale sparse graphs. These pre-defined structures allow each condensed node to encompass distinct neighborhoods. The variation in structure enables GNNs to capture broader information, leading to diverse node embeddings and enhanced performance. 
Additionally, GCSR explicitly models the inter-class correlations within the original graph structure and leverages them to reconstruct an interpretable condensed graph structure through the self-expressive property \cite{lu2012robust}. This approach transfers structural correlations from the original graph to the condensed graph, thereby facilitating better information preservation and enhancing the graph structure.

\noindent\textbf{Label augmentation.}
\textbf{PUMA} \cite{liu_puma_2023} and \textbf{CGC} \cite{gao2024rethinking} emphasize the significance of label quantity in GC and improve the condensation quality by incorporating pseudo labels from the original graph. The inclusion of additional supervision signals enhances class representation and ensures more accurate information preservation in the condensed graph.

\noindent\textbf{Feature augmentation.}
\textbf{GroC} \cite{li_attend_2023} designs a perturbation module in the process of generating condensed graphs. This module maximizes the distance between the condensed graph and the original graph by selectively perturbing the condensed node features that are insufficiently informative. 
By utilizing the adversarial training technique, these perturbations are capable of exploring the neighborhood areas within the parameter space of the condensed graph and significantly enhancing the gradient matching process.
\textbf{TinyGraph} \cite{liu2024tinygraph} aims to further compact the condensed graph by reducing the dimension of node features. It employs a graph attention network to compress the original graph features, with the compressed node features replacing the original ones as the condensation target during the condensation process.
By simultaneously optimizing the graph attention network and the condensed graph with a unified GC objective, TinyGraph effectively reduces the size of condensed graphs while preserving the downstream GNN performances.

\subsection{Generalized Graph Condensation}
\label{secmetgene}
The essence of generalization lies in preserving crucial task-related information while mitigating bias injection into the condensed graph. 
Nevertheless, some GNN relay models, employed for encoding both structure and feature information, inevitably result in a loss of structural information and create an entanglement between the condensed graph and the relay model.
These issues significantly affect the generalization ability to unseen models and tasks.
To explore the cause, \textbf{GDEM} \cite{liu_graph_2023} conducts an investigation across various GNN architectures, including different spatial and spectral GNNs.
They find that the performance gap increases when the relay model and downstream model have different architectures.
This is due to that distinct spectral preferences of GNNs have varying impacts on the significance of different graph eigenvectors \cite{bo2023survey,zhu2021interpreting}, consequently leading to alterations in node features.
For instance, if a GNN functions primarily as a low-pass filter, it will prioritize low-frequency information in node features, overlooking the full spectrum. The loss of high-frequency information will injure the high-pass filter GNNs' performance.

To address this issue, some generalized GC methods are developed from the spectral perspective.
For example, GDEM \cite{liu_graph_2023} eschews the conventional relay GNN and directly generates the eigenbasis for the condensed graph, thereby eliminating the spectrum bias inherent in relay GNNs.
To address the inconsistencies in eigenvector quantity and dimension, GDEM aligns the condensed eigenvectors with the most important eigenvectors of the original graph and matches the class features in the subspace defined by these eigenvectors. 
Apart from leveraging the advanced relay model, \textbf{SGDD} \cite{yang_does_2023} employs the Laplacian energy distribution (LED) to analyze the structural properties of the graph, quantifying the spectral shift between condensed and original graphs. 
A substantial LED shift suggests that crucial structural information from the original graph is lost, diminishing the generalization capabilities of the condensed graph.
Therefore, SGDD leverages the LED as a structural shift metric and designs an additional regularization to enhance the structure of the condensed graph. 
This regularization aligns the LED between the original graph and the condensed graph, facilitating the broadcast of structural information.
By preserving crucial structural information, SGDD effectively reduces the performance discrepancies observed across different GNN architectures and various tasks, e.g. node classification, anomaly detection and link prediction.

Besides these spectral methods, \textbf{OpenGC} \cite{gao2024graph} emphasizes the temporal generalization capacity of condensed graphs, particularly in dynamic open-world scenarios \cite{wu2020openwgl}. In this context, new nodes and classes are continuously integrated into the existing graph structure, resulting in a distribution shift compared to the original graph being condensed.
To adapt the condensed graph to these dynamic distribution shifts and manage graph data over an extended period, OpenGC employs temporal data augmentation to simulate evolving graph patterns and create multiple structure-aware environments. Furthermore, it incorporates invariance learning regularization in GC to preserve the invariant patterns across different temporal environments in the condensed graph.
By this means, OpenGC significantly enhances the adaptability of downstream GNNs by training them on the temporally generalized condensed graph, eliminating the need to laboriously design specific generalization modules for each GNN.

\subsection{Efficient Graph Condensation}
\label{secmeteffi}
The GC process requires significant computational resources.
To comprehend the root causes of heavy computation, we break down the GC process into three fundamental stages, including efficient graph encoding, efficient optimization, and efficient graph generation, and examine acceleration methods for each component.

\subsubsection{\textbf{Efficient Graph Encoding}} 
In GC, the graph encoding procedure is executed by the relay model, and it requires intensive computation primarily due to the neighbor explosion problem in the large original graph.
Generally, relay models used in GC methods adopt either the message-passing framework or the graph neural tangent kernel framework.
All these frameworks stack multiple propagation and transformation operations to aggregate and update information according to the graph structure. 
However, as multiple propagation processes are executed, the number of aggregated neighboring nodes increases exponentially, leading to significant computational demands.

To mitigate the expensive computation resulting from propagation, some GC methods borrow the idea from SGC \cite{wu2019simplifying} and remove the non-linear transformation layers between aggregation layers in the relay model. 
For example, \textbf{GCond} \cite{jin_graph_2022} and \textbf{PUMA} \cite{liu_puma_2023} utilize SGC to encode the original graph. \textbf{KiDD} \cite{xu_kernel_2023} simplifies the graph neural tangent kernel by removing non-linear activation at certain layers. By this means, the propagation of these methods can be pre-computed in the pre-processing procedure and only needs to be executed once. Instead of performing propagation during each training epoch, the online time complexity of GC is significantly reduced than utilizing other relay models.

\subsubsection{\textbf{Efficient Optimization}}
In GC, the optimization procedure requires numerous iterations to update the relay model before optimizing the condensed graph, leading to considerable computation.

\noindent\textbf{Gradient matching.}
To mitigate this problem, \textbf{DosCond} \cite{jin_condensing_2022} proposes the one-step gradient matching strategy which discards the iterated relay model update and only performs gradient matching for one single step. 
The theoretical analysis posits that minimizing the one-step matching loss serves a dual purpose: it not only directs the learning process of condensed graphs towards lower losses on original graphs but also delineates the trajectory for updating the condensed data.
Moreover, one-step matching eliminates the requirement of hyper-parameter tuning, such as the number of iterations of bi-level optimization, further simplifying the condensation process.
Besides simplifying the relay model updates in the inner loop optimization, \textbf{EXGC} \cite{fang2024exgc} aims to expedite the convergence of condensed graph generation in the outer loop. It begins by framing the bi-level optimization within an Expectation Maximization framework, then employs the Mean-Field variational approximation to update a subset of node features at once while maintaining complementary features fixed. Furthermore, EXGC observes that the condensation procedure follows a long-tail distribution, where a minority of condensed node features retain the majority of performance capabilities. However, this distribution leads to significant variability in the convergence speeds across different node sets. To address this, EXGC introduces post-hoc explanation methods to pinpoint the key subgraphs that hold most of the essential information for further convergence speedup.

\noindent\textbf{Kernel ridge regression.}
To avoid the bi-level optimization in GC, another research line involves {kernel ridge regression-based methods} \cite{xu_kernel_2023,wang_fast_2023,gao2024graph}. These methods reformulate the classification task as a kernel ridge regression task and incorporate the graph neural tangent kernel within kernel ridge regression to encode the original graph.
Kernel ridge regression provides an efficient closed-form solution, dispensing with the necessity for iterative computations in updating the relay model and resulting in faster convergence speeds.

\noindent\textbf{Distribution matching.}
Besides the gradient matching and kernel ridge regression, distribution matching changes the optimization principle towards approximating the original graph in the feature space, attracting significant attention in efficient GC studies. 
This approach leads to a shift in the role of the relay model within GC. It transitions from aligning task performances to transforming both original and condensed graphs into a unified space for distribution alignment. 
Consequently, distribution matching obviates the need to compute the gradients and trajectories w.r.t. parameters of the relay model in each training iteration, leading to a substantial reduction in computational complexity.
Specifically, \textbf{GCDM} \cite{liu_graph_2022}, \textbf{CaT} \cite{liu_cat_2023} and \textbf{PUMA} \cite{liu_puma_2023} compute node distributions for both original and condensed graphs in the same feature space and employ the maximum mean discrepancy \cite{gretton2012kernel} between each pair of class distributions as the optimization objective.
\textbf{GDEM} \cite{liu_graph_2023} decomposes the graph and matches the class distributions within sub-spaces formed by eigenvectors.
To enhance the fidelity of the condensation procedure, \textbf{DisCo} \cite{xiao2024disentangled} and \textbf{SimGC} \cite{xiao2024simple} introduce the pre-trained GNN in distribution matching and further align the node variations for performance enhancement.

\noindent\textbf{Clustering.}
Instead of updating the condensed graph via gradients, \textbf{CGC} \cite{gao2024rethinking} simplifies the distribution matching objective by recasting it as a class partition problem, which can be effectively resolved using any clustering method. 
By integrating this strategy with a predefined graph structure, CGC facilitates a training-free solution for generating condensed graphs. This innovation significantly accelerates the condensation process, enhancing the efficiency and scalability of GC.

\noindent\textbf{Computation tree preservation.}
Mirage \cite{gupta_mirage_2023} proposes a training-free solution for graph-level datasets. The underlying concept is that message-passing GNNs decompose the input graph into a multiset of computation trees according to the neighborhood aggregation. These computation trees capture the fundamental connectivity information in the graph and explicitly model the message-passing flows for each node. By preserving the most frequently occurring computation trees for each class, Mirage identifies the most crucial graph structures, thereby achieving a relay model-agnostic condensation method.

\subsubsection{\textbf{Efficient Graph Generation}}
Condensed graph generation targets two components: the feature matrix and the adjacency matrix. 
Modeling these matrices typically requires two distinct optimization processes. For example, \textbf{GCond} \cite{jin_graph_2022} models the condensed structure as a function of the condensed nodes, requiring the optimization of an additional adjacency matrix generator.
To simplify the optimization procedure, \textbf{MSGC} \cite{gao_multiple_2023} and \textbf{CGC} \cite{gao2024rethinking} leverage the pre-defined sparse graphs to eliminate the condensed graph optimization.
\textbf{GCSR} \cite{liu2024graph} incorporates the self-expressive property to construct the graph, enabling a closed-form solution for the calculation of the adjacency matrix.
Furthermore, several studies \cite{jin_graph_2022,zheng_structure_free_2023,liu_graph_2022,liu_cat_2023,liu_puma_2023} discard the adjacency matrix directly and instead use an identity matrix to represent condensed structure.
These structure-free GC methods implicitly encode the original graph’s topological structure into discriminative node features via relay GNNs and these enriched embeddings are informative to train different downstream GNNs.

\subsection{Fair Graph Condensation}
\label{secmetfair}
GC can effectively capture the essential task-specific characteristics of the original graph by exclusively optimizing for task performance.
However, this task-centric optimization approach may introduce fairness concerns, primarily attributed to the indiscriminate transmission of information into the condensed graph.
Especially in scenarios with intensive compression rates, GC tends to amplify biases as it prioritizes the preservation of predominant and representative information w.r.t the tasks. 
Consequently, GNNs trained on condensed graphs exhibit more severe fairness issues than those trained on original graphs \cite{feng_fair_2023,mao_gcare_2023} and challenge the applicability of GC in high-stake applications \cite{mehrabi2021survey,feng_fair_2023}.

To mitigate bias in condensed graphs, \textbf{FGD} \cite{feng_fair_2023} initially investigates the relationship between the feature spaces of the original graph and condensed graph, and constructs an estimator for sensitive features. Following this, the variance of the estimated sensitive group membership is employed as the bias measurement and integrated as the regularization to facilitate the unbiased condensed graph.
Moreover, \textbf{GCARe} \cite{mao_gcare_2023} directly regularizes the condensation process with adversarial training. It additionally incorporates a discriminator into the relay model, which is designed to predict the sensitive groups of node representations. 
The relay model operates as a generator, generating fair data to deceive the discriminator. 
Consequently, the condensation process is formulated as a min-max optimization problem, striking a balance between data generation and debiasing efforts.

While the two aforementioned methods address group fairness from distinct perspectives, their focus has primarily been on node classification tasks, leaving fairness concerns in other tasks largely unexplored. Furthermore, an empirical study by FGD has revealed that datasets also experience more pronounced issues with individual fairness when GNNs are trained on condensed graphs \cite{feng_fair_2023}. This underscores the need to broaden the scope of fairness research to include additional tasks and alternative fairness metrics, such as individual fairness \cite{song2022guide} and rank fairness \cite{chen2024fairness}.

\subsection{Robust Graph Condensation}
\label{secmetrobu}
GC typically assumes the availability of the clean original graph and emulates its distribution indiscriminately by the condensed graph.
However, this idealized assumption contradicts the complexities of data processing in practical scenarios.
Real-world graphs are often marred by various forms of noise—ranging from data collection errors \cite{randall2014use} and outdated information \cite{tu2023deep} to inherent variability in dynamic environments \cite{kang2019robust, gao2021training}.
Additionally, recent research \cite{wu2024backdoor} has highlighted vulnerabilities of GC to backdoor attacks, where malicious attackers can subtly manipulate the condensed graph while maintaining its apparent quality.
These unexpected noises or deliberate attacks all significantly impact the representation of the condensed graph and carry over to GNNs, compromising the model prediction accuracy and the resilience of GC in real-world applications.

Despite its significant relevance, robust GC has not been extensively studied, with only a limited number of studies addressing this vital field. \textbf{RobGC} \cite{gao2024robgc} emerges as a pioneer by specifically targeting structural noises and incorporates a denoising procedure into the graph condensation process.
It employs the condensed graph as a denoising signal to provide an overview for the structure and node features of the original graph, facilitating the correction of inherent noises.
By alternately executing graph condensation and denoising procedures, the quality of both the original and condensed graphs is mutually improved, ensuring the condensed graph retains only the essential information from the original graph. 
Moreover, the unified optimization in RobGC enables the denoising of unseen test graphs using the noise-free condensed graph, leading to more accurate inference results.

As a preliminary exploration, RobGC effectively demonstrates the feasibility and crucial need for developing robust GC methods. However, its application is currently limited to addressing structural noises. Expanding the research scope to include challenges such as node feature noise and label noise is imperative for enhancing the utility of GC across a broader spectrum of practical applications.

\begin{table*}[t]
\centering
\caption{Summary of structure modeling and sparsification methods in existing graph condensation research, with representative literature listed.
}
\begin{tabular}{l|l|l|l}
\toprule
\textbf{}                                 & \textbf{Methods}                                & \textbf{Principles}              & \textbf{Literature}                                                            \\ \midrule
\multirow{6}{*}{Structure modeling}       & Generative model                       & Node similarity         & GCond \cite{jin_graph_2022}, SGDD \cite{yang_does_2023}, DisCo \cite{xiao2024disentangled} \\ \cline{2-4} 
                                          & \multirow{3}{*}{Parameterization}      & \begin{tabular}[c]{@{}l@{}}Node self-expression \&\\ Inter-class structure correlation\end{tabular} & GCSR \cite{liu2024graph}                             \\ \cline{3-4} 
                                          &                                        & Low rank decomposition  & KiDD \cite{xu_kernel_2023}     \\ \cline{3-4} 
                      &                                        & Spectral decomposition  & GDEM \cite{liu_graph_2023}                         \\ \cline{2-4} 
                                          & \multirow{2}{*}{Pre-defined structure} & Node similarity         & MSGC \cite{gao_multiple_2023}, CGC \cite{gao2024rethinking}    \\ \cline{3-4} 
                                          &                                        & Structure-free method         & GCond \cite{jin_graph_2022}, SFGC \cite{zheng_structure_free_2023}                                          \\ \hline
\multirow{3}{*}{Structure sparsification} & Threshold                              &   -                   & GCond \cite{jin_graph_2022}                      \\ \cline{2-4} 
                                          & Regularization                         &   -                   & SGDD \cite{yang_does_2023}                         \\ \cline{2-4} 
                                          & Gumbel softmax                         &   -                   & DosCond \cite{jin_condensing_2022}, KiDD \cite{xu_kernel_2023}    \\ \bottomrule
\end{tabular}
\label{tab:adj}
\end{table*}

\section{Condensed Graph Generation}
\label{secgene}
Different from traditional graph reduction methods that trim the original graph, GC focuses on creating an entirely new condensed graph. 
An effective graph generation method not only accelerates optimization convergence but also enhances overall performance. In this section, we discuss the condensed graph generation methods from two aspects: feature and structure generation.

\subsection{Feature Generation}
\label{secgeneatt}

The feature matrix $\mathbf{X}'$ dominates parameter quantity in the condensed graph and the most of methods update it by gradients.
As a result, the feature matrix is significantly influenced by diverse initializations, potentially leading to varied convergence speeds.
To expedite the optimization process for the feature matrix, researchers have investigated a range of initialization methods as substitutes for conventional random initialization.
For instance, GCond \cite{jin_graph_2022} initiates the feature matrix by randomly selecting original nodes with corresponding labels. 
SFGC \cite{zheng_structure_free_2023} leverages the coreset method \cite{sener2017active} and chooses the representative nodes for initialization. 
On the other hand, MCond \cite{gao_graph_2023} and CTRL \cite{CTRL} demonstrate that employing the representative embeddings can accelerate the convergence of optimization. Accordingly, these methods first cluster original graph nodes based on the pre-defined compression rate, and then utilize the cluster centroids as the initialization of the feature matrix.

In addition to gradient-based optimization, which relies on different initializations, CGC \cite{gao2024rethinking} proposes generating the feature matrix through a closed-form solution. 
This approach eliminates the need for iterative updates of the feature matrix, thus ensuring a more precise and efficient feature generation process.

\subsection{Structure Generation}
\label{secgenestru}

The adjacency matrix $\mathbf{A}'$ of the condensed graph reflects relationships among condensed nodes and preserves the topological information of the original graph. The construction method of the adjacency matrix affects its characteristics, such as homophily, spectral properties, and sparsity, which in turn impact overall performance. As shown in Table \ref{tab:adj}, current studies employ three kinds of strategies for generating adjacency matrices: generative model, parameterization, and predefined structure.

\noindent\textbf{Generative model.}
A straightforward generation method involves employing the generative model to construct an adjacency matrix explicitly. 
For example, GCond \cite{jin_graph_2022} constructs the adjacency matrix based on homophily and generates the adjacency matrix by node similarities.
The generative model is designed to investigate the correlations between pairs of condensed nodes.
SGDD \cite{yang_does_2023} utilizes noise, along with the features and labels of the condensed graph, as inputs to the generative model for creating the adjacency matrix. 
DisCo \cite{xiao2024disentangled} transfers the link prediction model of the original graph to the condensed nodes, generating the corresponding edges for the condensed graph.
Although this method offers greater flexibility in controlling the generated structures, it suffers from substantial modeling complexity: the number of parameters in $\mathbf{A}'$ grows quadratically with the increase of condensed nodes.

\noindent\textbf{Parameterization.}
To mitigate the modeling complexity of graph structure, several GC methods propose to parameterize the adjacency matrix. 
GCSR \cite{liu2024graph} explores the self-expressive property among condensed nodes and further enhances the condensed graph by transferring the inter-class correlations from the original graph structure.
Besides, KiDD \cite{xu_kernel_2023} directly optimizes the low-rank decomposed matrices of the adjacency matrix to speed up the convergence. 
To preserve the spectral characteristics, GDEM \cite{liu_graph_2023} decomposes the adjacency matrix into an eigenbasis and constructs the condensed structure using the eigenvectors matched with the most significant eigenvectors of the original graph.

\noindent\textbf{Pre-defined structure.}
To further simplify the modeling of condensed graphs, several studies leverage pre-defined structures for the condensed graph.
For instance, MSGC \cite{gao_multiple_2023} pre-defines multiple sparse adjacency matrices for condensed nodes. By this means, MSGC reduces the complexity of the adjacency matrix while capturing a more diverse range of structural information compared to the single dense matrix.
Moreover, several GC methods have incorporated the structure-free strategy \cite{jin_graph_2022,zheng_structure_free_2023,liu_graph_2022,liu_cat_2023,liu_puma_2023} by employing a fixed identity matrix as the adjacency matrix. This strategy leads to a significant simplification of the computational requirements for the adjacency matrix.

\begin{table*}[]
\centering
\renewcommand{\arraystretch}{1.3}
\caption{The accuracy (\%) comparison between GC methods under different compression rates. ``OOM'' means out-of-memory. The highest accuracies are highlighted in bold, and runners-up are underlined. ``GM'' represents gradient matching. ``DM'' denotes distribution matching. ``KRR'' is kernel ridge regression, and ``TM'' represents trajectory matching. ``Original graph'' denotes the GNN performance which is trained on the original graph.}
\resizebox{\textwidth}{!}
{
\begin{tabular}{l|c|cccc|ccc|cc|cc|c}
\hline
\multirow{2}{*}{Dataset}    & \multirow{2}{*}{$r$} & \multicolumn{4}{c|}{GM}                                           & \multicolumn{3}{c|}{DM}                          & \multicolumn{2}{c|}{KRR}           & \multicolumn{2}{c|}{TM}               & \multirow{2}{*}{\begin{tabular}[c]{@{}c@{}}Origianl\\ graph\end{tabular}} \\ \cline{3-13}
                            &                      & GCond             & GCond-X        & DosCond  & SGDD              & GCDM     & GCDM-X            & SimGC             &  {\scriptsize GC-SNTK}        & {\scriptsize GC-SNTK-X}         & SFGC              & GCSR              &                                                                           \\ \hline
\multirow{3}{*}{Cora}       & 1.30\%               & 79.8±1.3          & 75.9±1.2       & 80.5±0.1 & 80.1±0.7          & 69.4±1.3 & 81.3±0.4          & 80.8±2.3          & {\ul 81.7±0.7} & \textbf{82.2±0.3} & 80.1±0.4          & 79.9±0.7          & \multirow{3}{*}{81.2±0.2}                                                 \\
                            & 2.60\%               & 80.1±0.6          & 75.7±0.9       & 80.1±0.5 & 80.6±0.8          & 77.2±0.4 & 81.4±0.1          & 80.9±2.6          & 81.5±0.7       & \textbf{82.4±0.5} & {\ul 81.7±0.5}    & 80.6±0.8          &                                                                           \\
                            & 5.20\%               & 79.3±0.3          & 76.0±0.9       & 80.3±0.4 & 80.4±1.6          & 79.4±0.1 & \textbf{82.5±0.3} & {\ul 82.1±1.3}    & 81.3±0.2       & {\ul 82.1±0.1}    & 81.6±0.8          & 81.2±0.9          &                                                                           \\ \hline
\multirow{3}{*}{Citeseer}   & 0.90\%               & 70.5±1.2          & {\ul 71.4±0.8} & 71.0±0.2 & 69.5±0.4          & 62.0±0.1 & 69.0±0.5          & \textbf{73.8±2.5} & 66.4±1.0       & 69.9±0.4          & {\ul 71.4±0.5}    & 70.2±1.1          & \multirow{3}{*}{71.7±0.1}                                                 \\
                            & 1.80\%               & 70.6±0.9          & 69.8±1.1       & 71.2±0.2 & 70.2±0.8          & 69.5±1.1 & 71.9±0.5          & {\ul 72.2±0.5}    & 68.4±1.1       & 69.9±0.5          & \textbf{72.4±0.4} & 71.7±0.9          &                                                                           \\
                            & 3.60\%               & 69.8±1.4          & 69.4±1.4       & 70.7±0.1 & 70.3±1.7          & 69.8±0.2 & {\ul 72.8±0.6}    & 71.1±2.8          & 69.8±0.8       & 69.1±0.4          & 70.6±0.7          & \textbf{74.0±0.4} &                                                                           \\ \hline
\multirow{3}{*}{Ogbn-arxiv} & 0.05\%               & 59.2±1.1          & 61.3±0.5       & 62.1±0.3 & 60.8±1.3          & 59.3±0.3 & 61.0±0.1          & 63.6±0.8          & {\ul 64.4±0.2} & 63.9±0.3          & \textbf{65.5±0.7} & 60.6±1.1          & \multirow{3}{*}{71.4±0.1}                                                 \\
                            & 0.25\%               & 63.2±0.3          & 64.2±0.4       & 63.5±0.1 & 65.8±1.2          & 59.6±0.4 & 61.2±0.1          & \textbf{66.4±0.3} & 65.1±0.8       & 65.5±0.1          & {\ul 66.1±0.4}    & 65.4±0.8          &                                                                           \\
                            & 0.50\%               & 64.0±0.4          & 63.1±0.5       & 63.7±0.2 & 66.3±0.7          & 62.4±0.1 & 62.5±0.1          & \textbf{66.8±0.4} & 65.4±0.5       & 65.7±0.4          & \textbf{66.8±0.4} & 65.9±0.6          &                                                                           \\ \hline
\multirow{3}{*}{Flickr}     & 0.10\%               & 46.5±0.4          & 45.9±0.1       & 46.0±0.3 & \textbf{46.9±0.1} & 46.1±0.1 & 46.0±0.1          & 45.3±0.7          & {\ul 46.7±0.1} & 46.6±0.3          & 46.6±0.2          & 46.6±0.3          & \multirow{3}{*}{47.2±0.1}                                                 \\
                            & 0.50\%               & \textbf{47.1±0.1} & 45.0±0.2       & 46.2±0.2 & \textbf{47.1±0.3} & 46.8±0.1 & 45.6±0.1          & 45.6±0.4          & 46.8±0.1       & 46.7±0.1          & 47.0±0.1          & 46.6±0.2          &                                                                           \\
                            & 1.00\%               & \textbf{47.1±0.1} & 45.0±0.1       & 46.1±0.1 & \textbf{47.1±0.1} & 46.7±0.1 & 45.4±0.3          & 43.8±1.5          & 46.5±0.2       & 46.6±0.2          & \textbf{47.1±0.1} & 46.8±0.2          &                                                                           \\ \hline
\multirow{3}{*}{Reddit}     & 0.05\%               & 88.0±1.8          & 88.4±0.4       & 89.8±0.1 & \textbf{90.5±2.1} & 89.3±0.1 & 86.5±0.2          & 89.6±0.6          & OOM            & OOM               & 89.7±0.2          & \textbf{90.5±0.2} & \multirow{3}{*}{93.9±0.0}                                                 \\
                            & 0.10\%               & 89.6±0.7          & 89.3±0.1       & 90.5±0.1 & \textbf{91.8±1.9} & 89.7±0.2 & 87.2±0.1          & 90.6±0.3          & OOM            & OOM               & 90.0±0.3          & {\ul 91.2±0.2}    &                                                                           \\
                            & 0.20\%               & 90.1±0.5          & 88.8±0.4       & 91.1±0.1 &{\ul 91.6±1.8}& 90.2±0.4 & 88.8±0.1          & 91.4±0.2          & OOM            & OOM               & 89.9±0.4          & \textbf{92.2±0.1} &                                                                           \\ \hline
\end{tabular}}
\label{tab:expeffective}
\end{table*}

\noindent\textbf{Condensed graph sparsification.}
The structure construction methods typically yield dense adjacency matrices with continuous values. These dense matrices often contain small but storage-intensive values that have minimal impact on GNN aggregation performance. Moreover, certain application scenarios, such as molecular graphs, require binary and discrete adjacency matrices for downstream tasks. Hence, it becomes essential to sparsify the generated adjacency matrices, and numerous sparsification methods have been proposed to tackle these challenges. For instance, GCond \cite{jin_graph_2022} utilizes the threshold to remove the entries with small values and justifies that suitable choices of threshold do not degrade performance but increase the scalability of the condensed graph.
SGDD \cite{yang_does_2023} employs the L2 norm of the adjacency matrix as a regularization term for sparsity control.
To construct binary and discrete adjacency matrices, DosCond \cite{jin_condensing_2022} and KiDD \cite{xu_kernel_2023} model each pair of nodes as an independent Bernoulli variable. Then, the Gumbel-Max reparametrization trick \cite{maddison2016concrete} is utilized to address the non-differentiability challenges in the optimization procedure.
\section{Empirical Studies}
Considering the diversity of optimization strategies and GC criteria, it is crucial to understand the characteristics of different GC methods and design appropriate modules for enhancing specific task performance.
Although this is not a paper centered on experiments and analysis, in this section we empirically analyze existing GC methods with diverse optimization strategies in terms of the five proposed criteria to provide the guidelines for both researchers and practitioners.

\subsection{Evaluation Metric}

Each criterion for the GC method is evaluated by a specific metric.

Effectiveness is assessed by comparing the accuracy of GNNs trained on condensed graphs at varying compression rates to those trained on the original graph. Methods that closely match the accuracy of the original setup are considered more effective.

Generalization is measured by comparing the average performance of models trained on condensed graphs across various GNN architectures and a range of downstream tasks, such as node classification, link prediction, anomaly detection, and temporal prediction. For a specific task $t$, the average performance of $n$ GNN architectures is defined as:
\begin{equation}
\label{eq_metric_gener}
p_t=\frac{1}{n} \sum_{i=1}^{n}acc_{i},
\end{equation}
where $acc_{i}$ represents the accuracy of the $i^{th}$ GNN architecture. A higher value indicates superior generalization ability for task $t$.

Efficiency is evaluated by the total time required for the condensation procedure, with shorter times indicating better practical usability.

Fairness is assessed by comparing the bias performance of GNNs trained on condensed graphs to those trained on the original graph. For the sensitive feature $s\in\{0, 1\}$, it is desired that GNN predictions remain independent of the sensitive feature \cite{feng_fair_2023,dai2021say}. With binary labels denoted as $y\in\{0, 1\}$, and model predictions as $\hat{y}\in\{0, 1\}$, model bias is quantified by demographic parity ($\Delta_{DP}$) \cite{beutel2017data} and equal opportunity ($\Delta_{EO}$) \cite{louizos2015variational}:
\begin{equation}
\small
\label{eq_metric_f1}
\Delta_{DP}=\left | P(\hat{y}=1|s=0)-P(\hat{y}=1|s=1) \right |,
\end{equation}
\begin{equation}
\small
\label{eq_metric_f2}
\Delta_{EO}=\left | P(\hat{y}=1|{y}=1, s=0)-P(\hat{y}=1|{y}=1,s=1) \right |.
\end{equation}
GC methods that effectively mitigate the bias of GNNs trained on the original graph are considered fairer.

Robustness is evaluated by the accuracy of GNNs trained on condensed graphs derived from original graphs with varying noise levels. Methods that achieve higher accuracy are considered to preserve core information more effectively and demonstrate greater robustness against noise in the original graph.

\begin{table}[]
\renewcommand{\arraystretch}{1.3}
\centering
\caption{The generalizability comparison of GC methods on Ogbn-arxiv dataset. ``Opt.'' and ``Avg.'' indicate the optimization strategy and average value, respectively. The highest accuracies are highlighted in bold, and runners-up are underlined. $r$ is set as 0.25\%.}
\resizebox{\linewidth}{!}
{
\begin{tabular}{l|l|cccccc|c}
\hline
\begin{tabular}[c]{@{}l@{}}Opt.\end{tabular} & Method    & SGC           & GCN           & SAGE          & APPNP         & Cheby         & GAT           & Avg.          \\ \hline
\multirow{4}{*}{GM}                                             & GCond     & 63.7          & 63.2          & 62.6          & 63.4          & 54.9          & 60.0          & 61.3          \\
                                                                & GCond-X   & 64.7          & 64.2          & 64.4          & 61.5          & 59.5          & 60.1          & 62.4          \\
                                                                & DosCond   & 63.3          & 63.5          & 62.1          & 63.5          & 55.1          & 60.4          & 61.3          \\
                                                                & SGDD      & 64.3          & 65.8          & 63.4          & 63.3          & 55.9          & 61.4          & 62.4          \\ \hline
\multirow{3}{*}{DM}                                             & GCDM      & 61.2          & 59.6          & 61.1          & 62.8          & 55.4          & 61.2          & 60.2          \\
                                                                & GCDM-X    & 64.4          & 61.2          & 63.4          & 60.5          & {\ul 60.2}    & 60.0          & 61.6          \\
                                                                & SimGC     & 64.3          & \textbf{66.4} & 60.4          & 61.5          & 54.7          & 61.1          & 61.4          \\ \hline
\multirow{2}{*}{KRR}                                            & {\scriptsize GC-SNTK}   & 62.7          & 65.1          & 62.9          & 62.6          & 55.1          & 61.8          & 61.7          \\
                                                                & {\scriptsize GC-SNTK-X} & 64.0          & 65.5          & 62.4          & 61.8          & 58.7          & 60.9          & 62.2          \\ \hline
\multirow{2}{*}{TM}                                             & SFGC      & {\ul 64.8}    & {\ul 66.1}    & {\ul 64.8}    & {\ul 63.9}    & \textbf{60.7} & \textbf{65.7} & \textbf{64.3} \\
                                                                & GCSR      & \textbf{65.6} & 65.4          & \textbf{65.4} & \textbf{64.4} & 58.9          & {\ul 63.5}    & {\ul 63.9}    \\ \hline
\end{tabular}}
\label{tab:gener}
\end{table}

\subsection{Experimental Settings}

\noindent\textbf{Methods.} To assess the efficacy of various optimization strategies, we evaluate 11 representative GC methods categorized under 4 distinct optimization strategies:
(1) gradient matching (GM)-based method: GCond, GCond-X~\cite{jin_graph_2022}, DosCond~\cite{jin_condensing_2022} and SGDD~\cite{yang_does_2023};
(2) distribution matching (DM)-based method: GCDM, GCDM-X~\cite{liu_graph_2022} and SimGC~\cite{xiao2024simple};
(3) KRR-based method: GC-SNTK and GC-SNTK-X~\cite{wang_fast_2023}; 
(4) trajectory matching (TM)-based method: SFGC~\cite{zheng_structure_free_2023} and GCSR~\cite{liu2024graph}. Notice that the suffix ``-X'' represents the graphless variant.

\noindent\textbf{Datasets.} Given the diversity of evaluation criteria, GC methods are evaluated across various datasets with distinct characteristics, including three transductive datasets (Cora, Citeseer~\cite{DBLP:conf/iclr/KipfW17} and Ogbn-arxiv~\cite{hu2020ogb}), two inductive datasets (Flickr and Reddit~\cite{DBLP:conf/iclr/ZengZSKP20}) and a dataset with sensitive feature (Pokec-n~\cite{dai2021say}).

\noindent\textbf{Implementations.}
Following GCond~\cite{jin_graph_2022}, we evaluate three compression rates ($r$) for each dataset to evaluate the effectiveness.
In the transductive setting, $N$ represents the original graph size, while in the inductive setting, $N$ indicates the sub-graph size observed in the training stage.
Two-layer GNNs with 256 hidden units are used for evaluation.

\noindent\textbf{Hyper-parameters.} 
The hyper-parameters are determined through the grid search on the validation set. We use the ADAM optimization algorithm to train all the models. The learning rate for the condensation process is determined through a search over the set \{0.01, 0.001, 0.0001\}. The weight decay is 5e-4. Dropout is searched from [0, 1).

\noindent\textbf{Computing Infrastructure.} The codes are written in Python 3.9 and Pytorch 1.12.1. The operating system is Ubuntu 18.0, and all models are trained on GPUs. 
All experiments are conducted on a server with Intel(R) Xeon(R) CPUs (Gold 6128 @ 3.40GHz) and NVIDIA GeForce RTX 2080 Ti 11GB GPUs.

\subsection{Experimental Results}

\noindent\textbf{Effectiveness.} The condensed graphs generated by GC methods are evaluated to train the same 2-layer GCN \cite{DBLP:conf/iclr/KipfW17} for node classification and the test accuracies with standard deviation are reported in Table \ref{tab:expeffective}. In the table, ``Original graph'' refers to the GCN performance which is trained on the original graph and we make the following observations.
(1) GC methods are not sensitive to the compression rate and a larger condensed graph size does not strictly indicate better performance. 
Even with extremely small compression rates, GC methods can achieve promising performance.
(2) The graphless variants demonstrate comparable performance to the method with the structure modeling, indicating the graph structure information can be effectively encoded into the condensed nodes.
(3) When comparing different optimization strategies, trajectory matching-based methods consistently show superior performance, outperforming other strategies across various datasets.
While KRR-based methods achieve competitive performance, the graph kernel in KRR is memory-intensive and does not scale well with large-scale graphs.
Distribution matching-based methods perform well on smaller datasets, whereas gradient matching-based methods are more effective on larger datasets.

\begin{figure}[t]
\centering
\includegraphics[width=0.45\textwidth]{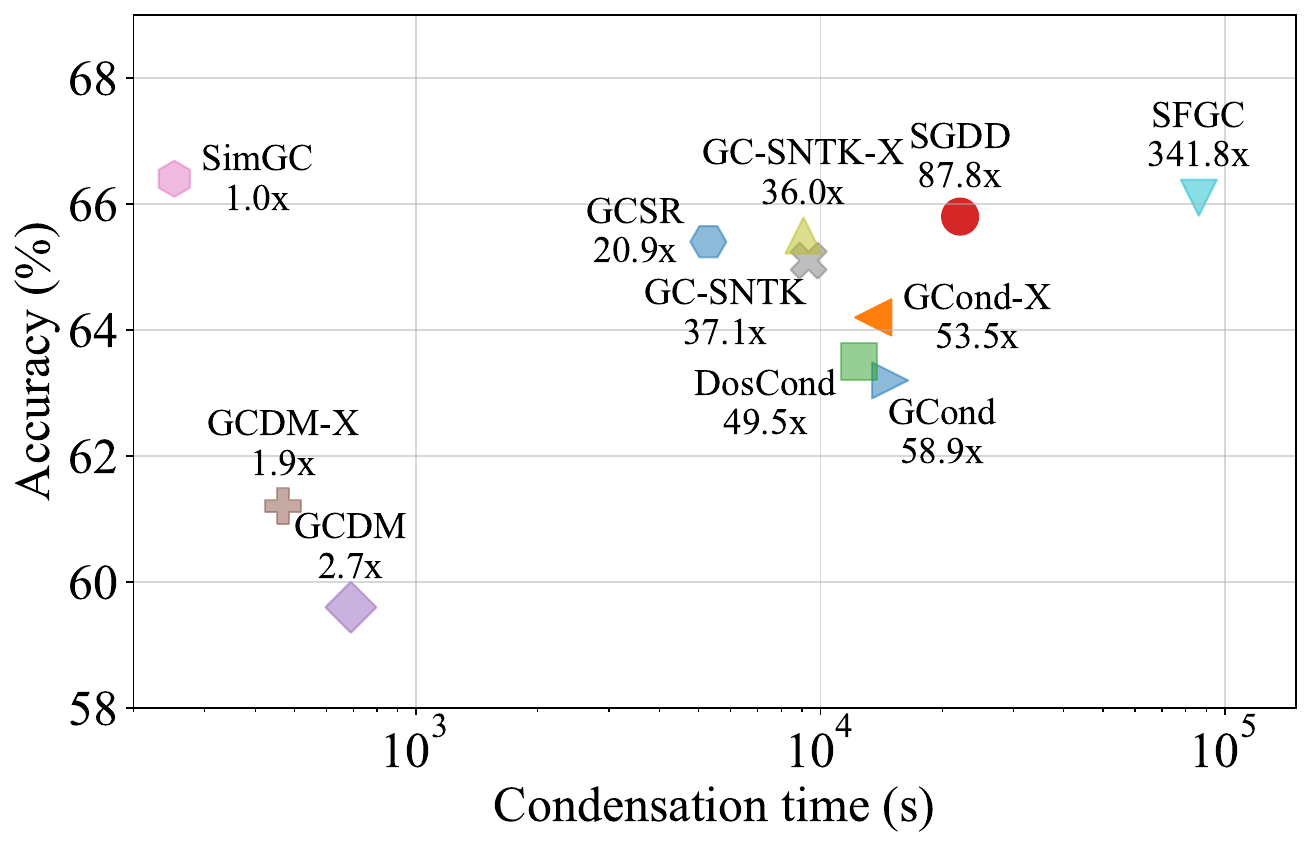}
\caption{The condensation time (seconds) comparison of different GC methods on Ogbn-arxiv dataset. $r$ is set as 0.25\%.} 
\label{fig:time}
\end{figure}

\noindent\textbf{Generalization.}
We adhere to the settings commonly used in most GC studies and assess the generalization of GC methods across various GNN models for the node classification task. The GNN models evaluated include GCN~\cite{DBLP:conf/iclr/KipfW17}, SGC~\cite{wu2019simplifying}, SAGE~\cite{hamilton2017inductive}, APPNP~\cite{gasteiger_predict_2019}, Cheby~\cite{defferrard2016convolutional} and GAT~\cite{DBLP:conf/iclr/VelickovicCCRLB18}.
The detailed accuracies are shown in Table \ref{tab:gener}.
Across various GNN architectures, most GC methods consistently achieve the best performance on GCN and SGC. This superior performance is attributed to the same convolution kernels \cite{jin_graph_2022} in the relay models of these GC methods, which share the same spectral characteristics \cite{liu_graph_2023} as those in GCN and SGC.
Among different optimization strategies, trajectory matching-based methods demonstrate a significant improvement over other methods, underscoring the effectiveness of trajectory matching in capturing essential information from the original graph and its superior generalization capabilities.

\noindent\textbf{Efficiency.}
We report the condensation times of various GC methods in Fig. \ref{fig:time} and make the following observations.
(1) Distribution matching achieves the most efficient condensation process among the optimization strategies, primarily due to its elimination of the need for gradient calculations. In contrast, trajectory matching and gradient matching are computational intensive.
(2) Without graph structure modeling, graphless variant methods expedite the condensation process.
(3) The one-step matching strategy used in DosCond eliminates the intensive model updates required in the inner loop while maintaining competitive classification performance.

\noindent\textbf{Fairness.}
Following FairGNN \cite{dai2021say} and FGD \cite{feng_fair_2023}, we evaluate the representative GC method in each optimization strategy on the Pokec-n dataset, and assess group fairness issues within a binary sensitive feature setting. 
As shown in Table \ref{tab:fair}, GC methods amplify the graph data bias to varying degrees. 
Gradient matching and distribution matching not only achieve the lowest node classification accuracy but also exhibit the most pronounced bias. 
More advanced optimization strategies, i.e., KRR and trajectory matching, can effectively reduce this bias while preserving predictive performance.

\begin{table*}[]
\renewcommand{\arraystretch}{1.3}
\centering
\caption{The comparison of AUC (\%), demographic parity ($\Delta_{DP}$, \%) and equal opportunity ($\Delta_{EO}$, \%) for GC methods with various optimization strategies.}
\resizebox{0.7\textwidth}{!}
{
\begin{tabular}{l|cc|cc|c|cc|c}
\hline
Opt.                      & \multicolumn{2}{c|}{GM} & \multicolumn{2}{c|}{DM} & KRR      & \multicolumn{2}{c|}{TM}              & \multirow{2}{*}{\begin{tabular}[c]{@{}c@{}}Origianl\\ graph\end{tabular}} \\ \cline{1-8}
Method                    & GCond      & SGDD       & GCDM       & SimGC      & GC-SNTK  & SFGC              & GCSR             &                                                                           \\ \hline
AUC $\uparrow$            & 52.2±1.9   & 55.1±0.9   & 53.0±1.6   & 55.3±1.3   & 54.4±1.1 & \textbf{55.5±1.6} & 54.9±0.6         & 72.9±0.2                                                                  \\
$\Delta_{DP} \downarrow $ & 3.0±1.2    & 2.5±1.0    & 2.3±1.4    & 2.4±1.1    & 2.0±1.7  & 2.1±1.2           & \textbf{1.9±0.9} & 0.2±0.1                                                                   \\
$\Delta_{EO} \downarrow $ & 3.2±1.0    & 3.0±1.1    & 2.7±1.5    & 2.8±1.2    & 2.6±1.6  & 2.2±1.1           & \textbf{2.0±1.2} & 1.4±1.6                                                                   \\ \hline
\end{tabular}}
\label{tab:fair}
\end{table*}

\noindent\textbf{Robustness.}
We evaluate the robustness of GC methods to structural noise, which involves randomly adding and removing edges. The noise level is set at 100\%, defined by the proportion of modified edges in the original graph \cite{jin2020graph}. 
As illustrated in Table \ref{tab:exprob}, the noisy graph structure significantly impacts the classification performance of models trained on condensed graphs. The performance disparity between noisy condensed graphs and noisy original graphs is greater than that observed with clean graphs, suggesting that structural noise severely affects the quality of condensed graphs.
Among the GC methods, distribution matching is most sensitive to random noises, while trajectory matching demonstrates robust performances.

\begin{table*}[]
\renewcommand{\arraystretch}{1.3}
\centering
\caption{The performance of GC methods against random structure noise on Ogbn-arxiv. Noise level is set as 100\%. ``Original graph'' indicates that GNNs are trained on the noisy original graph. In contrast, other methods involve training GNNs on the condensed graph.}
\resizebox{0.75\textwidth}{!}
{
\begin{tabular}{l|cc|cc|c|cc|c}
\hline
\multirow{2}{*}{Method} & \multicolumn{2}{c|}{GM} & \multicolumn{2}{c|}{DM} & KRR      & \multicolumn{2}{c|}{TM} & \multirow{2}{*}{\begin{tabular}[c]{@{}c@{}}Origianl\\ graph\end{tabular}} \\ \cline{2-8}
                        & GCond      & SGDD       & GCDM       & SimGC      & GC-SNTK  & SFGC       & GCSR       &                                                                           \\ \hline
Clean graph             & 63.2±0.3   & 65.8±1.2   & 59.6±0.4   & 66.4±0.3   & 65.1±0.8 & 66.1±0.4   & 65.4±0.8   & 71.4±0.1                                                                  \\
Noisy graph             & 50.1±0.5   & 52.9±0.9   & 49.4±0.7   & 51.7±0.4   & 52.7±0.7 & 54.6±0.7   & 55.5±0.5   & 61.1±0.1                                                                  \\ \hline
\end{tabular}}
\label{tab:exprob}
\end{table*}
\section{Applications and Resources}
Due to the efficiency in training GNN models and the capacity to conserve storage costs, GC contains vast application prospects and is increasingly being utilized across a variety of fields.
In this section, we first enumerate the diverse applications of GC, including hyperparameter/neural architecture search, graph continual learning, federated learning, inference acceleration, recommender systems, and heterogeneous graphs. 
Consequently, this section details the open-source libraries for GC methods, facilitating the advancement of future research in this domain.

\subsection{Practical Applications}
\subsubsection{\textbf{Hyper-Parameter/Neural Architecture Search}}
The hyper-parameter/neural architecture search \cite{gao2021graph} involves systematic experimentation with a range of hyper-parameters or architectures to identify the optimal combination for downstream tasks. This process typically requires the training of multiple models with varying hyper-parameters/architectures on the same dataset, incurring substantial computational costs.
To mitigate the computational burden, \textbf{HCDC} \cite{ding_faster_2022} introduces GC in this problem and additionally generates the synthetic validation data by matching the hyper-parameter gradients.
This ensures that the performance ranking of hyper-parameters/architectures on condensed datasets is consistent with that on the original datasets. By integrating GC, HCDC significantly reduces the time required for the search process.

\subsubsection{\textbf{Graph Continual Learning}}
Graph continual learning \cite{febrinanto2023graph} is the approach that focuses on the continuous acquisition of knowledge from dynamic and evolving graph data.
In this realm, a critical challenge is the catastrophic forgetting problem, where a model's performance on previously learned data deteriorates significantly when trained on new data.
To mitigate this issue, \textbf{CaT} \cite{liu_cat_2023} and \textbf{PUMA} \cite{liu_puma_2023} incorporate GC into the process of graph continual learning, leveraging the condensed graphs to preserve historical information within a memory bank. To maintain a balance between historical and newly incoming data, CaT strategically trains the model exclusively on the condensed graph.
This method not only addresses the challenge of data imbalance in graph continual learning but also significantly enhances learning efficiency.

\subsubsection{\textbf{Federated Learning}}
Federated learning \cite{fu2022federated} is a distributed framework that safeguards private data, involving multiple clients with exclusive data and a central server. In this setting, clients train local models with their unique data and then transmit the parameters to the server, which aggregates and redistributes the updated knowledge. A key challenge in federated learning is the non-i.i.d nature of client data, leading to biased local models and slow convergence, increasing the communication load between the server and clients. To address this, \textbf{FedGKD} \cite{pan_fedgkd_2023} integrates GC at the client level to efficiently extract the local task features, enhancing the similarity computation and ensuring better incorporation of local information. Combined with global aggregation, FedGKD significantly improves collaboration results in federated learning environments.
Besides integrating GC at the client level, \textbf{FedGC} \cite{yan2024federated} explores the federated graph condensation task, which aims to learn a unified condensed graph at the server level, encompassing knowledge from various clients.
By matching the aggregated gradients from clients, FedGC tackles the data heterogeneity problem and generates a global condensed graph, facilitating efficient model training in the cross-silo scenarios.

\subsubsection{\textbf{Inference Acceleration}}
In practical graph systems, new present nodes will be connected with existing nodes in the original graph, then the message-passing process is executed to aggregate information and generate representations for inference. 
However, the extensive size of large original graphs poses a considerable challenge for real-time inference \cite{gao2023accelerating}. 
This difficulty primarily stems from the ``neighbor explosion'' problem, where the number of aggregated nodes exponentially increases with the propagation depth.
To reduce the high computational demands and latency inherent in GNNs, \textbf{MCond} \cite{gao_graph_2023} adopts GC and proposes to learn a mapping from original nodes to condensed nodes. This approach allows direct information propagation on the small condensed graph, resulting in a significant acceleration of the inference process.

\subsubsection{\textbf{Recommender Systems}}
Recommender systems are widely used in various industries such as e-commerce and social media, aiming to suggest relevant items or content to users. In a typical recommender system, the user-item interactions can be represented as a bipartite graph. As the number of users and items grows, this graph becomes increasingly large and complex, posing significant computational challenges. To handle this issue, \textbf{Distill-CF} \cite{sachdeva2022infinite} applies GC to recommender systems. It adopts a multi-step Gumbel sampling trick to condense the original user base into a limited number of representative users, which are then iteratively updated through distilling knowledge from the original graph. By replacing the original user-item graph with the synthesized small graph for training and inference, this method leads to more scalable and efficient recommender systems.

\subsubsection{\textbf{Heterogeneous Graphs}}

Heterogeneous graphs \cite{wang2022survey,gao2023semantic} are adept at modeling complex systems characterized by multi-typed entities and relations, thereby providing a broad range of practical applications and significant potential for generalization.
However, their complex structure and inherent heterogeneity present significant challenges in data management and in training heterogeneous GNNs. 
In light of these challenges, \textbf{HGCond} \cite{gao2024heterogeneous} extends GCond \cite{jin_graph_2022} from simple graphs to heterogeneous graphs, generating a condensed heterogeneous graph for efficient model training. 
To address the heterogeneity problem, HGCond initializes condensed nodes by clustering same-type original nodes and further enhances the gradient matching optimization by constraining the parameter space of the relay model with orthogonal parameter sequences. 
This approach significantly enhances the scalability and generalization of heterogeneous GNNs on heterogeneous graphs, thereby broadening their applications in managing complex, large-scale systems.

\subsection{Open-Source Libraries}

Several notable PyTorch-based GC libraries have been developed based on the widely-used GNN libraries PyTorch Geometric (PyG)\footnote{\href{https://www.pyg.org/}{https://www.pyg.org/}} and Deep Graph Library (DGL)\footnote{\href{https://www.dgl.ai/}{https://www.dgl.ai/}}. These libraries provide essential tools and benchmarks to advance research and applications in graph-based systems.

\begin{itemize}
\item 
{\textbf{GCondenser}}\footnote{\href{https://github.com/superallen13/GCondenser}{https://github.com/superallen13/GCondenser}} \cite{liu2024gcondenser}, compatible with both PyG and DGL, supports 6 GC methods evaluated across 6 diverse graph datasets for node classification tasks. It also includes implementations for applying GC methods to graph continual learning and hyperparamter sweeping by Optuna \cite{akiba2019optuna}.
\item 
{GC-\textbf{Bench}}\footnote{\href{https://github.com/RingBDStack/GC-Bench}{https://github.com/RingBDStack/GC-Bench}} \cite{sun2024gc} is a unified GC benchmark based on PyG. It incorporates 9 GC methods assessed over 12 graph datasets for node- and graph-level tasks, including node/graph classification, link prediction, node clustering, and anomaly detection.
\item 
{\textbf{GraphSlim}}\footnote{\href{https://github.com/Emory-Melody/GraphSlim}{https://github.com/Emory-Melody/GraphSlim}} \cite{gong2024gc} is a PyG-based library for graph reduction, including sparsification, coarsening, and condensation. It evaluates 7 GC methods on 5 graph datasets for node classification and extends the utility of GC to neural architecture search. Additionally, it assesses the robustness of GC methods by integrating the DeepRobust \cite{li2020deeprobust} library.
\end{itemize}
\section{Challenges and Future Directions}
Although current methodologies in GC have shown remarkable performance across various applications, the field still grapples with inherent limitations and unresolved challenges. In this section, we summarize the current challenges and identify several future research directions worth exploring.

\subsection{Condensation for Diverse Graphs} 
As an innovative approach for graph reduction, current GC methods have predominantly focused on simple graphs. However, the graph data in practical applications is far more complex, showcasing a wide spectrum of characteristics including heterophilic graphs \cite{zheng2022graph}, digraphs \cite{bang2008digraphs}, and dynamic graphs \cite{kazemi2020representation}, etc. Each type of graph brings forth its own unique features, necessitating a deep understanding of their structures and a requirement for diverse GC strategies. Such a comprehensive grasp is not only critical for effectively managing the increasing volumes of data but also plays a key role in widening the horizons of their applications.

\subsection{Task-Agnostic Graph Condensation} 
The optimization strategies currently employed in GC are specifically designed for the classification task, heavily relying on class labels. 
However, the entanglement of the GC optimization process with task-related labels can potentially hinder performance in other downstream tasks.
The recent study, CTGC \cite{gao2024contrastive}, investigates self-supervised learning to enhance the task adaptability of GC for link prediction, and clustering tasks. This confirms the feasibility of extending GC to a wider spectrum of practical scenarios and delineates promising directions for future research, including anomaly detection \cite{akoglu2015graph}, graph reconstruction \cite{xia2021graph}, and graph generation \cite{zhu2022survey}.

\subsection{Efficient Evaluation for Condensed Graphs} 
In the realm of GC, the primary objective is to create high-quality condensed graphs. However, a systematic and well-established methodology for assessing the condensed graph quality is notably absent in the field \cite{zheng_structure_free_2023}. This gap is particularly critical given the complex optimization involved in GC. The assessment of the condensed graph quality is essential not only for monitoring the condensation progress but also for potentially streamlining the optimization through early-stop strategies. Currently, GC practices leverage the downstream models, i.e., GNN \cite{jin_graph_2022} or graph neural tangent kernel \cite{zheng_structure_free_2023}, as the indicator for assessment. However, this method adds considerable computational overhead to the GC process. Moreover, the concept of quality in condensed graphs is inherently complex, varying significantly across different applications and necessitating a multifaceted approach to evaluation. Considering these challenges, there is a pressing requirement to develop comprehensive and efficient evaluation metrics.  Such metrics would not only accelerate the condensation process but also broaden the range of GC applications.

\subsection{Secure Graph Condensation} 
The condensation of graphs not only facilitates efficient training of GNNs but also contributes significantly to reducing the storage overhead and accelerating data transmission. 
However, the security of highly compressed graph datasets to various privacy threats \cite{sun2022adversarial} remains under-explored. 
The recent study \cite{wu2024backdoor} has exposed vulnerabilities of GC and injected malicious information into condensed graphs to backdoor the GNNs trained on them. This highlights concerns over additional threats, such as adversarial attacks \cite{gao2024robgc} and inference breaches \cite{hu2022membership}, which pose significant challenges for deploying GC in security-sensitive domains, including confidential communications, financial systems, and industrial controls.
In light of these challenges, the development of secure, privacy-preserving GC algorithms becomes imperative.

\subsection{Explainable Graph Condensation} 
Another promising avenue for GC research is the development of explainable GC methods. 
A relevant work \cite{fang2024exgc} utilizes explainable algorithms to assess the importance of condensed nodes and prune the condensation redundancy. However, this method still provides limited insight into how condensed graphs represent the original data \cite{yuan2022explainability}.  
An explainable condensation can illuminate the intricate patterns within the graph, thereby enhancing user trust and acceptance. This is particularly important in sectors like healthcare \cite{wang2020recent} and finance \cite{wang2021review}, where the interpretability and justification of decision-making processes are critical. 
Consequently, by elucidating the logic behind data transformations, explainable GC has the potential to open up new avenues for its application, establishing GC as a more reliable and transparent tool in data-centric technology fields.
\section{Conclusion}
This survey presents an up-to-date and comprehensive overview of GC and classifies the existing research into five distinct categories aligned with the proposed essential GC evaluation criteria. It further explores two critical components of GC: optimization strategies and condensed graph generation, facilitating a deep understanding of GC techniques. Additionally, it highlights the current challenges and emerging perspectives in GC, with the objective of inspiring and guiding future research in this evolving field. Overall, this survey seeks to serve as a beacon for researchers and practitioners, guiding them through the evolving landscape of GC and fostering advancements that could reshape the use of graph data in various practical applications.

\bibliography{IEEEabrv,ref}
\bibliographystyle{IEEEtran}




\end{document}